\documentclass[draft]{agujournal2019}
\usepackage{url} %this package should fix any errors with URLs in refs.
\usepackage{lineno}
\usepackage[inline]{trackchanges} %for better track changes. finalnew option will compile document with changes incorporated.
\usepackage{soul}
\usepackage{amsmath}
\usepackage[normalem]{ulem}
\usepackage{xcolor}
\usepackage{graphicx}
\usepackage{subcaption} 
\usepackage{adjustbox}
\usepackage{amsfonts} % RSA

\draftfalse

\journalname{JGR: Machine Learning and Computation}

\begin{document}

%%%%%%%%%%%%%%%%%%%%%%%%%%%%%%%%%%%%%%%%%%%%%%%
%  TITLE
%
% (A title should be specific, informative, and brief. Use
% abbreviations only if they are defined in the abstract. Titles that
% start with general keywords then specific terms are optimized in
% searches)
%
%%%%%%%%%%%%%%%%%%%%%%%%%%%%%%%%%%%%%%%%%%%%%%%

% Example: \title{This is a test title}

\title{Toward generative machine learning for boosting ensembles of climate simulations}

%%%%%%%%%%%%%%%%%%%%%%%%%%%%%%%%%%%%%%%%%%%%%%%
%
%  AUTHORS AND AFFILIATIONS
%
%%%%%%%%%%%%%%%%%%%%%%%%%%%%%%%%%%%%%%%%%%%%%%%

% Authors are individuals who have significantly contributed to the
% research and preparation of the article. Group authors are allowed, if
% each author in the group is separately identified in an appendix.)

% List authors by first name or initial followed by last name and
% separated by commas. Use \affil{} to number affiliations, and
% \thanks{} for author notes.
% Additional author notes should be indicated with \thanks{} (for
% example, for current addresses).

\authors{Parsa Gooya\affil{1},Reinel Sospedra-Alfonso\affil{1}, and Johannes Exenberger\affil{2,3}}

\affiliation{1}{Canadian Centre for Climate Modeling and Analysis, Environment and Climate Change Canada, Victoria, British Columbia, Canada}
\affiliation{2}{Vienna University of Technology}
\affiliation{3}{Institute of Software Technology and Artificial Intelligence, Graz University of Technology}

% Corresponding author mailing address and e-mail address:

% Example: \correspondingauthor{First and Last Name}{email@address.edu}

\correspondingauthor{Parsa Gooya}{parsa.gooya@ec.gc.ca}

%%%%%%%%%%%%%%%%%%%%%%%%%%%%%%%%%%%%%%%%%%%%%%%
% KEY POINTS
%%%%%%%%%%%%%%%%%%%%%%%%%%%%%%%%%%%%%%%%%%%%%%%
%  List up to three key points (at least one is required)
%  Key Points summarize the main points and conclusions of the article
%  Each must be 140 characters or fewer with no special characters or punctuation and must be complete sentences

% Example:
% \begin{keypoints}
% \item	List up to three key points (at least one is required)
% \item	Key Points summarize the main points and conclusions of the article
% \item	Each must be 140 characters or fewer with no special characters or punctuation and must be complete sentences
% \end{keypoints}

\begin{keypoints}
\item A framework for conditional variational autoencoders is presented to generate large ensembles of climate data from limited training sample.
\item Variational autoencoders learn the underlying distribution of data and generate ensembles with extremes beyond the limited training sample.
\item Proper representation of output noise is key to avoid overly smooth outputs, spectral bias and loss of variability.
\end{keypoints}

%%%%%%%%%%%%%%%%%%%%%%%%%%%%%%%%%%%%%%%%%%%%%%%
%
%  ABSTRACT and PLAIN LANGUAGE SUMMARY
%
% A good Abstract will begin with a short description of the problem
% being addressed, briefly describe the new data or analyses, then
% briefly states the main conclusion(s) and how they are supported and
% uncertainties.

% The Plain Language Summary should be written for a broad audience,
% including journalists and the science-interested public, that will not have 
% a background in your field.
%
% A Plain Language Summary is required in GRL, JGR: Planets, JGR: Biogeosciences,
% JGR: Oceans, G-Cubed, Reviews of Geophysics, and JAMES.
% see http://sharingscience.agu.org/creating-plain-language-summary/)
%
%%%%%%%%%%%%%%%%%%%%%%%%%%%%%%%%%%%%%%%%%%%%%%%

%% \begin{abstract} starts the second page

\begin{abstract}

% The quantification of uncertainty due to irreducible climate internal variability is key for decision making and the assessment of extremes. Such  uncertainties are typically quantified with simulation ensembles produced with physics-based climate models. However, due to computational constrains, there is typically a trade-off between producing larger ensembles needed for accurate uncertainty estimation and increasing model resolution to better resolve fine scale dynamics. Generative machine learning is an emergent tool to address those issues. 

Accurately quantifying uncertainty in predictions and projections arising from irreducible internal climate variability is critical for informed decision‑making. Such uncertainty is typically assessed using ensembles produced with physics‑based climate models. However, computational constraints impose a trade‑off between generating the large ensembles required for robust uncertainty estimation and increasing model resolution to better capture fine‑scale dynamics. Generative machine learning offers a promising pathway to alleviate these constraints. We develop a conditional Variational Autoencoder (cVAE) trained on a limited sample of climate simulations to generate arbitrary large ensembles. The approach is applied to output from monthly CMIP6 historical and future scenario experiments produced with the Canadian Centre for Climate Modelling and Analysis' (CCCma's) Earth system model CanESM5. We show that the cVAE model learns the underlying distribution of the data and generates physically consistent samples that reproduce realistic low- and high-moment statistics, including extremes. Compared with more sophisticated generative architectures, cVAEs offer a mathematically transparent, interpretable, and computationally efficient framework. Their simplicity lead to some limitations, such as overly smooth outputs, spectral bias, and underdispersion, that we discuss along with strategies to mitigate them. Specifically, we show that incorporating output noise improves  the representation of climate‑relevant multiscale variability, and we propose a simple method to achieve this. Finally, we show that cVAE-enhanced ensembles capture realistic global teleconnection patterns, even under climate conditions absent from the training data.

\end{abstract}

\section*{Plain Language Summary}

% The accurate projection of extremes and the quantification of uncertainty using physics-based climate models require large ensembles of highly resolved climate fields and are thus limited by computational resources. Recent advances in generative machine learning have proven successful in overcoming these limitations for weather prediction. However, this problem remains challenging for climate predictions due to the scarcity of training data and the non-stationarity imposed by a changing climate. We examine the ability of conditional Variational Autoencoders (cVAEs) to generate very large ensembles of  climate projections in a simplified framework. We train the VAE model using a single realization of CMIP6 historical and ssp245 simulations of Temperature at Surface (TAS) from CanESM5. We evaluate against the full ensemble to show the generalization skill of VAE models to learn the underlying distribution of the data allowing for the generation of climate fields with realistic spatial and temporal statistics consistent with the population.

We examine how generative machine learning methods can expand climate model ensembles used to quantify uncertainty arising from internal climate variability. Traditional ensemble approaches rely on computationally expensive physics‑based climate models, creating a trade‑off between ensemble size and model resolution. To address this, we train a conditional Variational Autoencoder (cVAE) on a limited set of monthly historical and scenario simulations from the Canadian Earth system model CanESM5. The cVAE learns the statistical structure of the training data and generates samples that preserve key physical characteristics, including realistic distributions, extremes, and large-scale spatial patterns. Although cVAEs exhibit limitations such as spectral bias and underdispersion, we analyze these behaviors and outline practical mitigation strategies. We demonstrate that adding output noise improves the representation of multiscale climate variability. The resulting cVAE-enhanced ensemble reproduces credible global climate relationships, even under climate conditions absent from the training data.

%%%%%%%%%%%%%%%%%%%%%%%%%%%%%%%%%%%%%%%%%%%%%%%
%
%  BODY TEXT
%
%%%%%%%%%%%%%%%%%%%%%%%%%%%%%%%%%%%%%%%%%%%%%%%

%%% Suggested section heads:
\section{Introduction}

% The climate system is chaotic, implying that small errors in the initial conditions of climate simulations grow with time, making projections inherently uncertain \cite{slingopalmer2011}. Uncertainty quantification in climate modeling is therefore crucial for assessment of risk and decision making \cite{mankin2020,tebaldi2005quantifying}. Climate simulations rely on ensembles of model runs with each member generated using slightly different initial conditions sampling a range of possible initial climate states \cite{slingopalmer2011, Houtekamer1995}. The resulting spread over the ensemble projection provides an estimation of uncertainty. The more well-sampled the ensemble distribution is, the more reliably will the ensemble spread cover the range of realizable climate states at a given forcing \cite{Gooya-2023, Lehner-2020, Leutbecher2019}. The size of the climate model ensemble is limited by computational resources, making it necessary for researchers to balance the trade-off between the spatial resolution of the simulations and the ensemble size \cite{Sospedra-2021, Swart-2019}. 

The climate system is inherently chaotic, meaning that small perturbations of climate states amplify over time and make both seasonal-to-decadal (S2D) predictions and long-term projections fundamentally uncertain \cite{MLclimate2024, diffensemblegen2024}. Quantifying this uncertainty is essential for risk assessment and informed decision-making in climate applications \cite{sacco:hal-03685523}.  To represent uncertainty, climate modeling typically relies on simulation ensembles, with each member of the ensemble representing a possible realization of the climate system. For S2D predictions (spanning a few months to ten years), ensemble members are initialized with slightly perturbed initial states to sample a range of plausible climate trajectories driven by predictable modes of internal variability, such as El Ni\~no-Southern Oscillation (ENSO), and by external forcing, such as greenhouse gasses and volcanic aerosols \cite{merryfield2020,bkm13,sb19}. For long-term projections (spanning several years to over a century), simulations are instead initialized from sufficiently different climate states to span slow-evolving internal variability, with predictability arising only from the imposed external forcing. In this context, the spread across the ensemble provides an estimate of long-term uncertainty, and broader, more thoroughly sampled ensembles offer a more reliable representation of the range of realizable climate states under a given forcing \cite{Gooya-2023, Lehner-2020, Leutbecher2019}. Consequently, large ensembles are necessary for a robust estimation of the long-term uncertainty arising from both fast- and slow-evolving variability as well as to estimate the likelihood of low probability high impact extreme events. In practice, ensemble size is constrained by computational cost, imposing trade-offs between model resolution and the number of ensemble members that can be feasibly produced \cite{Sospedra-2021, Swart-2019}.

Recent advances in deep learning and generative machine learning have enabled the weather prediction community to substantially reduce the cost of ensemble forecasting  \cite{MLclimate2024, diffensemblegen2024}. These efforts range from improving prediction system output statistics, such as the spread of deterministic prediction ensembles \cite{sacco:hal-03685523} and probabilistic forecasts \cite{DLpost-processing}, to data-driven generation of high-dimensional weather-like samples from target forecast distributions \cite{diffensemblegen2024}. Generative models extract statistical priors that support conditional or unconditional sampling of learned probability distributions \cite{diffensemblegen2024}, increasing computational efficiency for tasks that traditionally required large computing resources with physics‑based models. For climate modeling, applications include probabilistic regional downscaling \cite{dynamicalgeneariveds2024, daust2024capturing}, simultaneous emulation and downscaling of climate models \cite{simultaneous2025}, and probabilistic climate model emulation \cite{cachay2024probabilistic, non-intrusice-correction-det, sorensen2024probabilisticframeworklearningnonintrusive, wang2025gen2generativepredictioncorrectionframework}, with generative ML emulators showing notable promise in the prediction of extreme event statistics by learning corrections to reduced complexity climate simulations such as outputs from a coarsely resolved numerical model \cite{sorensen2024probabilisticframeworklearningnonintrusive} or an auto-regressive Gaussian emulator \cite{wang2025gen2generativepredictioncorrectionframework}.

In this study, we employ a conditional Variational Autoencoder (cVAE) \cite{kingma2014, rezende2014} to generate arbitrary large samples of monthly near-surface air temperature (TAS) under historical or future scenario forcing. We are interested in learning the monthly TAS distribution of events over multi-decade time-scales consistent with the underlying climate. cVAEs are generative models that learn low dimensional representations of data and generate new data samples while conditioning on relevant information to control the generation process. They consist of an encoder network that maps the input data to a probability distribution over a latent space, often a multivariate Gaussian, and a decoder network that reconstructs the original data from samples drawn from this latent distribution \cite{prince2023understanding, dorta2018structureduncertaintypredictionnetworks}. Here, we condition both the encoder and decoder on a low rank representation of the input data that capture the warming trend to constrain and guide the distribution of the model’s output. This approach is suitable for climate ensemble generation tasks for which the data vary according to a potentially predictable component  (i.e., externally forced and, in the case of predictions, internally generated due to, e.g., ENSO) and a component that is not predictable (i.e., weather-induced noise or, in the case of long-term projections, all internally generated variability). The expectation is that cVAEs learn representations of  modes of variability conditioned on the dominant predictable components and generate unseen climate fields consistent with the learned modes. We chose cVAEs over more sophisticated generative models because of their straightforward and efficient training, their foundation in closed-form mathematical principles, and their greater interpretability. Our goal is not to design the most permanent model, but to show that inexpensive generative deep learning models trained on relatively small data samples can be used to produce ensembles of climate simulations with realistic behavior in both low- and high-order statistics, including the representation of extremes. Remarkably, we train our cVAE model on a \textit{single} realization of CCCma's Earth system model CanESM5 \cite{Swart-2019} and generate an arbitrary large ensemble of physically meaningful realizations. The generated ensemble is evaluated against the full ensemble of CanESM5 simulations that does not include the member used for training. Since the data used for training spans a sufficiently long period relative to the frequency of events assessed (e.g., ENSO events generated from 40-year training data), training on one member is sufficient for cVAE's performance. However, for a proper estimation of long-term uncertainty, training on more ensemble members and/or conditioning on the ensemble mean over two or more members is more appropriate. For predictions, on the other, a single member may suffice. We expand on this point in section 2.2.

% \textcolor{blue}{\textcolor{red}{ [Here, a couple of sentences are needed to justify the training on one ensemble member in the context of climate projections and historical simulations].[I suggest including them in the data section. I already added some arguments. Let me know what you think]}]}. Note that for a proper estimation of long-term uncertainty (i.e., uncertainty in the estimates of the externally-forced long-term trend) more ensemble members, are needed; for climate predictions, on the other hand, one member may suffice. If the data record is sufficiently long relative to the frequency of events that are assessed, then one member for projections can be enough to assess the skill on climate timescales. For example, when assessing ENSO for an ensemble generated from training over a 40 years period.

%Our simple approach and limited training data is justified by the seasonal to interannual variability... the representation of slow-varying modes would require more ensemble members... projections vs predictions...

The remainder of the paper is organized as follows. cVAEs are introduced in section \ref{subsec:generative_model}. Data and data preprocessing are discussed in section \ref{subsec:data_n_prepro}. The cVAE model and the training process are given in section \ref{subsec:model_n_training}, while details of the model inference are presented in section \ref{subsec:inference}, respectively. An important aspect of model inference is accounting for residual noise that is not learned during training. Section \ref{subsec:DN} examines the impact of including decoder noise during inference. Results and discussions are given in section \ref{sec:results_n_discussions}. Section \ref{sec:conclusions} ends with the conclusions.

\section{Data and Methods}
\label{sec:data_n_methods}

\subsection{Generative model}
\label{subsec:generative_model}

Conditional variational autoencoders (cVAEs) \cite{kingma2014, rezende2014, cVAE2015} are deep generative models that learn representations of high-dimensional distributions by formulating them as a ``variational inference" problem. A cVAE defines a joint distribution between the input data space $\mathcal{X}$ and an unobserved dimensionally-reduced latent space $\mathcal{Z}$ \cite{szwarcman2024, ghosh2020}, conditioned to $c$. The latent variable $z$ can be regarded as a lower-dimensional representation of ${x}$ coming from the conditional prior distribution $p({z|c})$. 

The generative process of cVAE consists of first generating a set of latent variables ${z} \in \mathcal{Z}$ from a prior distribution $p({z}|c)$, and then generating (sampling) the data ${x} \in \mathcal{X}$ according to the output distribution function $p_{\theta}({x} | {z}, c)$, usually parameterized as a Gaussian \cite{cVAE2015}. The distribution $p_{\theta}({x} | {z}, c)$, known as the generative model or a probabilistic $\textit{decoder}$, is derived using a neural network with parameters $\theta$. These are estimated through the maximum likelihood of the generated samples \cite{cVAE2015}, a challenging task that involves the intractable posterior $p({z} | {x}, c)$. Consequently, a distribution $q_{\phi}({z} | {x},c)$ known as the “recognition” model is introduced, often assumed to be a Gaussian, to approximate the true posterior $p({z} | {x}, c)$ \cite{kingma2014, cVAE2015}. The recognition model is derived using a separate neural network of parameters $\phi$, known as the probabilistic \textit{encoder}. 

Following \citeA{kingma2014}, \citeA{cVAE2015} showed that the parameters $\theta$ and $\phi$ for the cVAE can be efficiently estimated using the stochastic gradient variational Bayes (SGVB) framework, where the variational lower bound of the log-likelihood is used as a surrogate objective function:

\begin{align}
\label{eq:elbo}
\log p(x|c)
&= KL\left(q_{\phi}(z|x,c)\Vert p(z|x,c)\right)
 + \mathbb{E}_{q_{\phi}(z|x,c)}\left[
 -\log q_{\phi}(z|x,c) + \log p(x,z|c)
 \right] \notag \\
&\ge - KL\left(q_{\phi}(z|x,c)\Vert p(z|c)\right)
 + \mathbb{E}_{q_{\phi}(z|x,c)}\left[
 \log p_{\theta}(x|z,c)
 \right],
\end{align}

\noindent In Eq. \ref{eq:elbo}, the term ${KL}$ is the Kullback–Leibler divergence and the second term is the expectation of the log-likelihood of samples over the distribution $q_{\phi}({z} | {x},c)$ \cite{prince2023understanding}. Together, both the \textit{encoder} and \textit{decoder} models form the conditional variational autoencoder. The full derivation of Eq. \ref{eq:elbo} is provided in the supplements. 

%\noindent \textcolor{red}{\textbf{[NOT SURE WETHER THIS IS NEEDED HERE, OR AT ALL]}} While the prior distribution of the latent space of a cVAE is conditional, e.g. a normal distribution whose parameters depend on the condition, depending on the specific experiment and dataset, one could use the simplifying assumption that the latent space is condition independent (e.g. standard normal as for the standard VAE model) which is what we do here.  

To facilitate the computation of gradients with respect to the parameters of the encoder neural network, the reparameterization trick is used \cite{kingma2014}. Assuming a standard Gaussian distribution for the latent variables, the first term of Eq. \ref{eq:elbo} has a closed form, while the second term can be approximated by drawing samples ${z}^{(l)}$ ($l = 1, \dots, L$) from the Gaussian approximation to the posterior distribution (recognition model) $q_{\phi}({z} | {x},c)$. This reduces the objective function to:

% \begin{equation}
% \label{eq:loss}
%     \mathcal{L}({x}; \theta, \phi) = \frac{1}{2} \sum_{i=1}^{k} \left( 1 + \log \left( {\sigma_{i}}^2(x,c) \right) - {\mu_{i}(x,c)}^2 - {\sigma_{i}(x,c)}^2 \right) + \frac{1}{L} \sum_{l=1}^{L} \log p_{\theta}({x} | {z}^{(l)}, c),
% \end{equation}

\begin{equation}
\label{eq:loss}
    \mathcal{L}({x}; \theta, \phi) = \frac{1}{2} \sum_{i=1}^{k} \left( 1 + \log \left( {\sigma_{i}}^2(x,c) \right) - {\mu^2_{i}(x,c)} - {\sigma^2_{i}(x,c)} \right) + \frac{1}{L} \sum_{l=1}^{L} \log p_{\theta}({x} | {z}^{(l)}, c),
\end{equation}

\noindent which is to be maximized with respect to the parameters $\theta$ and $\phi$. Here ${\mu_{i}(x,c)}$ and ${\sigma_{i}(x,c)}$ are the means and standard deviations of $q_{\phi}({z} | {x},c)$ for the latent dimensions $i = 1, \dots, k$. We emphasize that $\theta$ and $\phi$ denote the parameters of the neural networks used to determine the decoder and encoder distributions, respectively, and are not to be mistaken for the parameters of the distributions. 

The cVAE model is thus formulated as follows:

\begin{equation}
\label{eq:cVAE}
    \begin{aligned}
        &\textit{Encoder:} \quad q_{\phi}({z} | {x}, {c}) = \mathcal{N} \left( {\mu}_{NN_{\phi}}({x}, {c}), {\sigma}_{NN_{\phi}}^2({x}, {c}) {I} \right) \\
        &\textit{Decoder:} \quad p_{\theta}({x} | {z}, {c}) = \mathcal{N} \left( {\mu}_{NN_{\theta}}({z}, {c}), {\sigma^2}{I} \right) \\
        % &\textit{Prior:} \quad p_{\omega}({z} | {c}) =   p({z}) = \mathcal{N} \left( {0}, {I} \right) \\
        &\textit{Prior:} \quad\quad p(z|c) = p(z) = \mathcal{N} \left( {0}, I \right) \\
    \end{aligned}
\end{equation}

\noindent where ${I}$ is the identity matrix and ${\sigma^2}$ is a location independent (constant) decoder noise. The encoder maps the input images $x\in\mathcal{X}$ into distributions over the latent space whose mean and standard deviation are output from the neural network and form $q_{\phi}({z} | {x},c)$, the Gaussian approximation to the posterior distribution. The decoder then maps the samples in the latent space back to the data space $\mathcal{X}$. Under this formulation, the second term in the cVAE loss function (Eq. \ref{eq:loss}) is proportional to the Mean Square Error (MSE) of the generated samples (see supplements), which encourages accurate reconstructions of data samples, and the KL divergence term regularizes a structured latent distribution. With this loss function, the cVAE learns the conditional distribution of the data through a lower dimensional latent representation that extracts information useful to generate data instances in a structured and continuous latent space. In the output space, the decoder models through ${\mu}_{NN_{\theta}}({z}, c)$ the important aspects of the data as explained by the latent variable ${z}$, and the remaining unmodeled aspects are ascribed to the noise ${\sigma^2}{I}$ \cite{prince2023understanding}. While the noise is usually assumed to be unstructured and independent of the data, we show that the decoder noise becomes especially important for climate data where multi-scale variability exists. In section \ref{subsec:inference}, we describe our approach to approximate the decoder noise and how it can relate to model failures, under-dispersivity, and loss of fine-scale variability. 

\subsection{Data and preprocessing}
\label{subsec:data_n_prepro}

We employ monthly values of near-surface air temperature (TAS) spanning 75 years (1951-2025), produced with the Canadian Earth System Model version 5 (CanESM5) \cite{Swart-2019} under CMIP6's historical and SSP2-4.5 scenario forcing \cite{eyering-2016}. The raw CanESM5 ensemble has 25 realizations of approximately 2.8$^{\circ}$ spatial resolution ($64\times128$ in lat$\times$lon), generated from different initial conditions that are produced from 50-year intervals of a piControl simulation. This would capture a range of realizable internal variability of the model's climate. From the 25-member ensemble, we select a single ensemble member for training, grossly under-sampling CanESM5's climate and its extreme events. The goal is to boost the ensemble size of our extremely limited training sample to generate TAS fields with distributions consistent with CanESM5's climate, including extremes. 

For this task, we train a cVAE model to reconstruct monthly TAS maps conditional on a compressed low dimensional embedding of the same input fields. For training, we use data from the full period (1950-2020) and reserve (2021-2025) for validation (Sec. 2.3 and S2). We standardize the input by removing the monthly climatology and dividing by the monthly standard deviation across all training years, at each grid cell. As a result, the cVAE model focuses on learning the interannual variability of the data conditioned on the state of the climate extracted through condition embedding. The compressed condition embedding only learns the dominant mode of variability in the data. Therefore, it is expected to learn a signal close to the dominant predictable components that are common to the generated ensemble. When training on a single realization, as we do here, this is reminiscent of seasonal-to-decadal predictions, as the conditioning learns the forced component and a phased-locked realization of internally generated variability (e.g., ENSO).  For climate projections, a conditioning that better isolates the forced component (e.g., average over several ensemble members, warming due to greenhouse gases ) would be desirable.  Given our time frame of analysis of the generated ensemble based on events that evolve on monthly to interannual time scales,  training on a single ensemble member suffices. 

\subsection{Model and training}
\label{subsec:model_n_training}

We use multilayer perceptrons (MLPs) for the encoder and decoder networks, with a latent variable $z \in \mathbb{R}^{k}, k=500$, and a standard Gaussian prior that is \textit{condition independent} (i.e. $p(z|c) = p(z) = \mathcal{N}(0,I)$). This is a reasonable assumption, as the goal is to learn the internal variability of the climate on interannual time scales. For condition embedding, we use a separate MLP denoted ${F}_{NN} ({x})$ that outputs an embedding vector $c \in \mathbb{R}^{2}$ passed to both the encoder and the decoder. Details on network architectures are given in the supplements. Importantly, in such limited data regime (840 data points), the simplicity of the model is important since overly parametrized models are prone to overfitting.

The choice of a much lower dimension for the condition embedding compared to the latent space is necessary for two reasons. First, we want the condition embedding network to extract only the dominant predictable signals rather than focusing on residual variability specific to the realization used for training. For example, Fig. S1 shows that the condition embedding learns mainly the long-term trend. Second, since the condition embedding is extracted from the same TAS field that is being reconstructed, if the embedding contains enough information to skillfully reconstruct the input, then the KL penalty encourages the network to ignore the latent space and only rely on the condition to minimize KL in the loss function. This is a phenomenon known as ``posterior collapse" \cite{betaannealing}, which implies that the network essentially collapses to a deterministic autoencoder. Therefore, a bottle-neck conditioning vector is necessary. 

Additionally, we employ beta annealing \cite{betaannealing} during training, a technique to linearly scale the weight of the KL divergence term in the loss function from 0 to 1 over the first 10 epochs during training. In this way, by slowly increasing the weight of the KL constraint in the latent space, the network starts as a simple autoencoder using the latent space together with the condition embedding for accurate reconstruction of TAS fields, which helps with learning a meaningful latent space. Posterior collapse where KL loss asymptotically approaches zero is then tracked from the training/validation KL divergence. We use batch size of 100 and a cosine decreasing learning rate with a starting maximum of 0.0001. The validation set is used to monitor posterior collapse and apply early stopping if the validation MSE does not improve over 15 epochs. Training the cVAE takes less than 10 minutes on a single GPU.

\subsection{Inference}
\label{subsec:inference}

Once trained, the model can be used to generate an arbitrarily large sample at each time instance using a conditioning field (i.e., a temperature map at a specific time), by sampling the latent space according to $p({z})$, decoding the samples with the conditional generative decoder, and sampling the decoder distribution (Eq. \ref{eq:cVAE}). This combined inference process takes a few seconds to run on a single GPU. We condition on TAS fields from a single ensemble member (same realization used for training) and generate 24-member ensembles to compare with the remainder 24 members out of the raw 25-member ensemble from CanESM5 (we refer to this ensemble as the population excluding train sample). We further apply a mean bias correction to the annual average of the generated TAS fields relative to the same conditioning input. The protocol for sample generation is described next. 

The $k$-dimensional normal prior $\mathcal{N}\left( {0}, I \right)$ in cVAE models is often chosen to simplify the latent sampling %${z}_m \in \mathbb{R}^{k}$ ($m = 1, \dots, M$) 
${z}^{(m)} \in \mathbb{R}^{k}$ ($m = 1, \dots, M$) at generation or inference time. During training, the $KL$ term in the loss function serves as a regularization constraint to encourage learning a latent space where samples fall under this normal distribution. While one can adjust the weight of the $KL$ term in the loss function (Eq. \ref{eq:elbo}), i.e. $(1+\beta) \, {KL}\left(q_{\phi}(z|x,c) \,\Vert\, p(z|c)\right)$ with $\beta\geq0$ \cite{cVAE2015}, latent samples do not always fall perfectly under the prior normal distribution \cite{distVAE2023} where they could be more spread out in one dimension or have small covariance between dimensions. This is particularly important at inference time when generating extremes and producing samples that reflect the correct tail probabilities \cite{IEEE}. To account for this, we could leverage the explicit formulation of the cVAE latent space and use the distribution of the encoded training data to guide inference sampling \cite{prince2023understanding, IEEE, AtmConvVAE2021, Taiwan}. For instance, scaling the standard deviation of the prior distribution at inference time has been shown to provide an efficient control for synthesis towards more extreme scenarios and a wider range of internal variability \cite{IEEE, gooya2025probabilisticbiasadjustmentseasonal}. Here, we approximate the Gaussian prior at inference time with the covariance matrix ${\Sigma}_{z}^{train}$ whose structure is acquired from the latent encodings of the training samples 
via the posterior $q_{\phi}({z} | {x},c)$,

\begin{eqnarray}
\label{eq:sigma_train_z}
{\Sigma}_{z}^{\text{train}} 
& = & 
\overline{\left( Z - \overline{Z} \right)\left( Z - \overline{Z} \right)^{\top}}
\nonumber \\
\end{eqnarray}

where

\begin{eqnarray}
Z 
& = & 
\begin{bmatrix}
z_1^\top \\
z_2^\top \\
\vdots \\
z_N^\top
\end{bmatrix},
\qquad
z_i \sim q_{\phi}\!\left( z \mid x_i, c_i \right) 
\end{eqnarray}
% \begin{eqnarray}
% \label{eq:sigma_train_z}
%          {\Sigma}_{z}^{train} & = & \bar{ ({z} - \bar{z}) (z - \bar{z})^\text{T}} \nonumber \\
%          z & \sim & q_{\phi}({z} | {x_{train}},c)         
% \end{eqnarray}

% \begin{equation}
% \label{eq:sigma_train_z}
%     \begin{aligned}
%          \quad  
%          {\Sigma}_{z}^{train} = \mathbb{E}_ {{x} \sim {\text{train}} } [ ({z} - \mathbb{E}[z]) (z - \mathbb{E}[z])^\text{T}], \quad    z \sim q_{\phi}({z} | {x_{}},c)
%     \end{aligned}
% \end{equation}
\noindent for $x_i,c_i$ pairs from the training dataset of size $N$. Therefore, at inference time we sample from the prior distribution $\mathcal{N}\left( {0}, {\Sigma}_{z}^{train} \right)$.

Each latent sample generated with this prior distribution is mapped into the output space with the decoder, following a Gaussian distribution with mean ${\mu}_{NN_{\theta}}(z,c)$ and variance given by the decoder noise (Eq. \ref{eq:cVAE}). The variability in the decoder output is (implicitly) taken as minimal when the mean is used as the output, which is often the case in regression based machine learning tasks. %(e.g.,\citeA{Improving_Seasonal_Forecast_2022, Taiwan}, etc.)}. 
For TAS data however, one can expect high frequency unpredictable variability (e.g., resulting from weather noise) superimposed to the variability explainable by  ${\mu}_{NN_{\theta}}(z,c)$ in the decoder, which the VAE model is likely to filter out through the bottle-neck latent variable. Therefore, representing such meaningful noise in the decoder on top of ${\mu}_{NN_{\theta}}(z)$ is important to avoid overly smooth fields. We note that the same could be true for any model trained with MSE as reconstruction loss, where the output converges to the mean of all possible learnable functions that vary depending on finite training data \cite{Goodfellow-deeplearning}. Now, modeling decoder variability with a constant diagonal matrix is based on the simplifying assumption that the noise is unstructured, random, and location-independent. While that might be the case for optimally-learned neural networks with sufficiently expressive architectures, this is not true in general. %In principle, one can relax this assumption and try learning location-dependent noise using another neural network.
\citeA{dorta2018structureduncertaintypredictionnetworks} argue that while the noise is commonly considered as unstructured inherent in the data, reconstruction errors (or residuals from the mean in the decoder Gaussian) are often caused by data deficiencies, limitations in the model capacity (e.g., suboptimal architecture) and suboptimal parameter estimation. Consequently, the decoder's variability is generally structured. 

% \citeA{dorta2018structureduncertaintypredictionnetworks} approximate the noise structure using a \textcolor{red}{multi-variate Gaussian} distribution with a sparse covariance matrix dependent on the data that is learned with another neural network \textcolor{red}{\textit{Decoder:} $\mathcal{N}\left( \mu_{NN_\theta}({z},c), {\Sigma}_{{NN}_{\theta}}({z},c) \right)$)} using the trained VAE model. 

In principle, one can relax the unstructured noise assumption and learn location-dependent noise using a neural network. For example, \citeA{dorta2018structureduncertaintypredictionnetworks} learn the noise structure using a multivariate Gaussian distribution with a sparse covariance matrix (\textit{Decoder:} $\mathcal{N}\left( \mu_{NN_\theta}({z},c), {\Sigma}_{{NN}_{\theta}}({z},c) \right)$) as a function of the data. We follow a simpler approach and estimate a covariance matrix from the residuals of the training data. Specifically, we estimate the covariance noise structure from the errors in the reconstruction of the training data. That is, we use the decoder distribution $ \mathcal{N} \left( {\mu}_{NN_{\theta}}({z},c), {\Sigma}^{train}_x \right)$ with

\begin{eqnarray}
\label{eq:sigma_train_x}
         {\Sigma}^{train}_x & = & \overline{({X}_{res} - \overline{X_{res}}) ({X}_{res} - \overline{{X}_{res}})^\text{T}} 
\end{eqnarray}

where

\begin{eqnarray}
X_{res} 
& = & 
\begin{bmatrix}
x_1 - {\mu}_{NN_\theta}({z_1},c_{1})  ^\top \\
x_2 - {\mu}_{NN_\theta}({z_2},c_{2})  ^\top\\
\vdots \\
x_N - {\mu}_{NN_\theta}({z_N},c_{N})  ^\top
\end{bmatrix},
\qquad
z_i \sim q_{\phi}\!\left( z \mid x_i, c_i \right) 
\end{eqnarray}

%  \begin{equation}
% \label{eq:sigma_train_x}
%     \begin{aligned}
%           \quad  
%          {\Sigma}_{\text{train}} = \mathbb{E}_ {{x}_{\text{train}} } [ ({x}_{\text{res}} - \mathbb{E}[{x}_{\text{res}}]) ({x}_{\text{res}} - \mathbb{E}[{x}_{\text{res}}])^\text{T}] \quad , \quad    {x}_{\text{res}} = {x}_{\text{train}}  - {\mu}_{NN_\theta}({z}|{x}_{\text{train}} ) \quad
%     \end{aligned}
% \end{equation}

We therefore update the cVAE in Eq. \ref{eq:cVAE} and proceed with the following model at inference time,

\begin{equation}
\label{eq:cVAE_updated}
    \begin{aligned}
        % &\textit{Encoder:} \qquad q_{\phi}({z} | {x}, {c}) = \mathcal{N} \left( {\mu}_{NN_{\phi}}({x}, {c}), {\sigma}_{NN_{\phi}}^2({x}, {c}) {I} \right) \\
        &\textit{Decoder:} \qquad p_{\theta}({x} | {z}, {c}) = \mathcal{N} \left( {\mu}_{NN_{\theta}}({z}, {c}), \Sigma^{train}_x  \right) \\
        % &\textit{Prior:} \quad p_{\omega}({z} | {c}) =   p({z}) = \mathcal{N} \left( {0}, {I} \right) \\
        &\textit{Prior:} \qquad\quad p(z|c) = p(z) = \mathcal{N} \left( {0}, \Sigma^{train}_z  \right) \\
        &\textit{Condition:} \quad c = {F}_{NN} ({x})
    \end{aligned}
\end{equation}

\noindent where $\Sigma^{train}_z$ and $\Sigma^{train}_z$ are given by Eqs. \ref{eq:sigma_train_z} and \ref{eq:sigma_train_x}, respectively, and $F_{NN}$ is the MLP model mapping the input space to $\mathbb{R}^2$.

% \textbf{\textcolor{red}{[Check. Perhaps make this link in a separate, brief subsection.]}} This can be regarded as analogous to online bias correction of climate projections in Earth System Models where climatological nudging tendencies acquired from an assimilation run are applied at prediction time to bring the model closer to the observed trajectory \cite{onlinebiascorrection}. While this covariance matrix approximation does not depend on the model generated output and could be rendered suboptimal, we show that it significantly improves the model performance, spectral bias, and its potential for underdispersivity. 

% \subsection{cVAE with and without decoder noise}
\subsection{Evaluation of cVAE with and without decoder noise}
\label{subsec:DN}

\begin{figure}
\centering

% ---------- Row 1 ----------
\begin{minipage}{0.48\textwidth}
    \centering
    \adjustbox{padding=0pt 0pt 0pt 25pt}{\includegraphics[width=\linewidth]{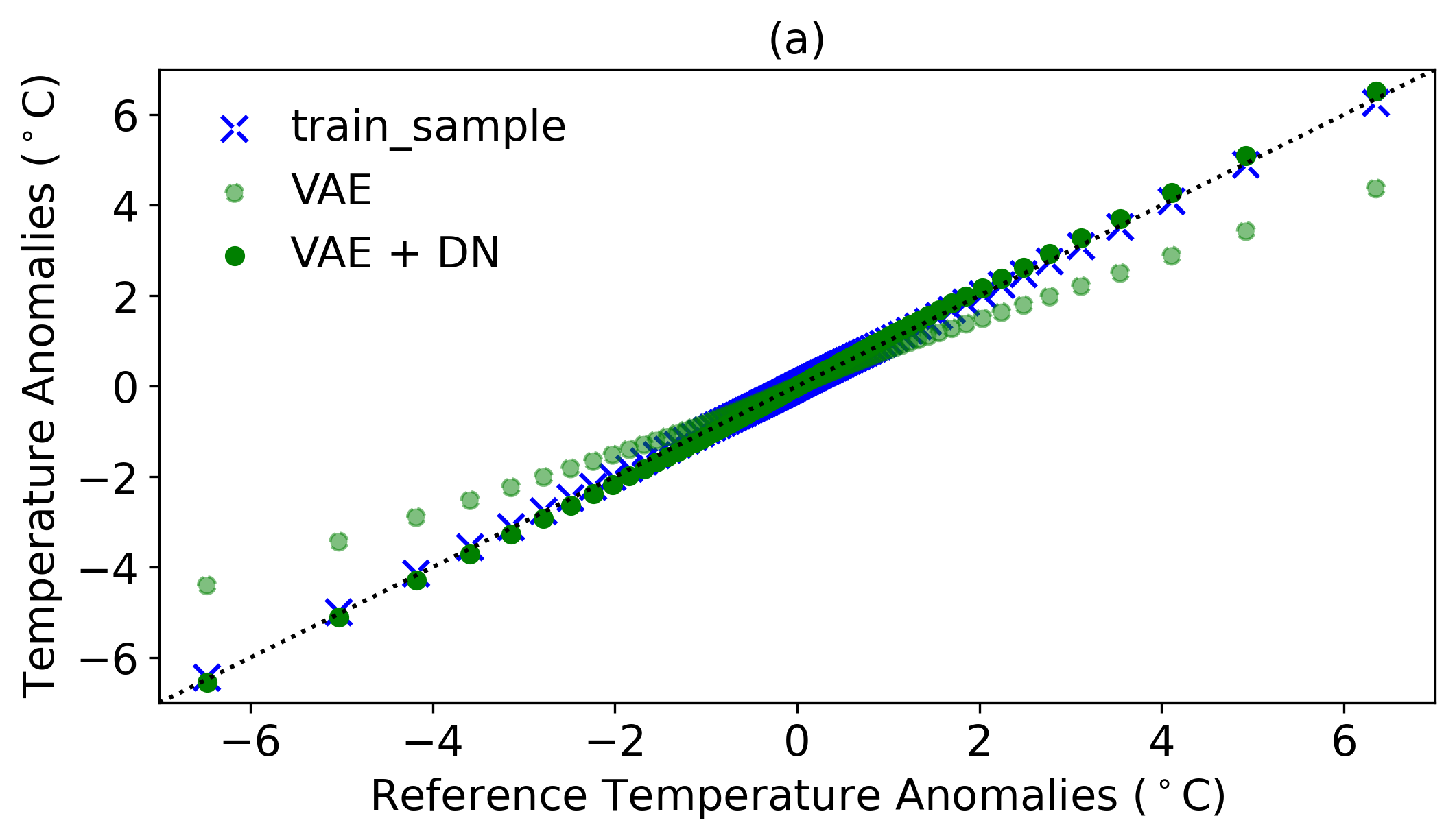}}
\end{minipage}\hfill
\begin{minipage}{0.48\textwidth}
    \centering
    \includegraphics[width=\linewidth]{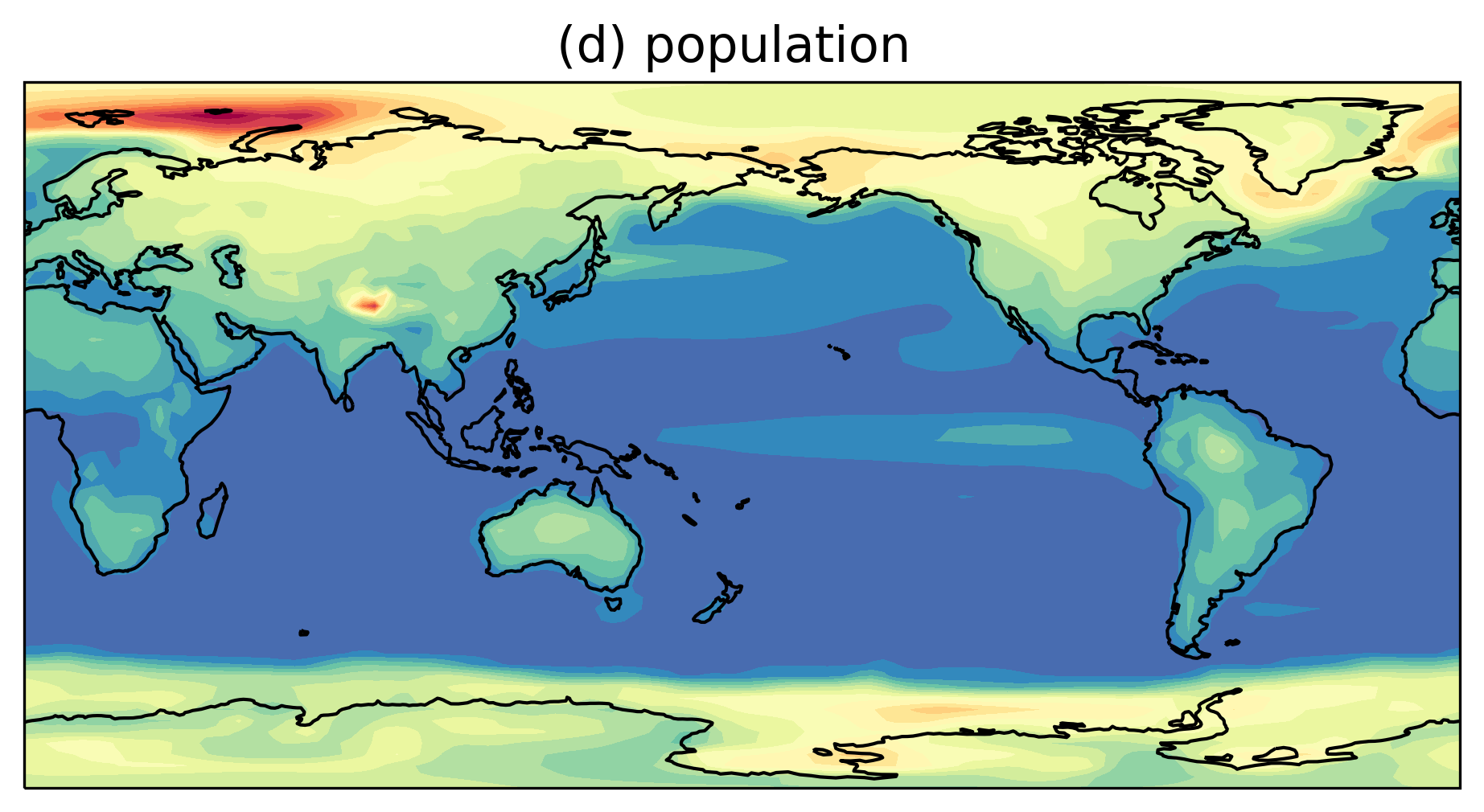}
\end{minipage}

\vspace{0em}

% ---------- Row 2 ----------
\begin{minipage}{0.48\textwidth}
    \centering
    \adjustbox{padding=0pt 0pt 0pt 25pt}{\includegraphics[width=\linewidth]{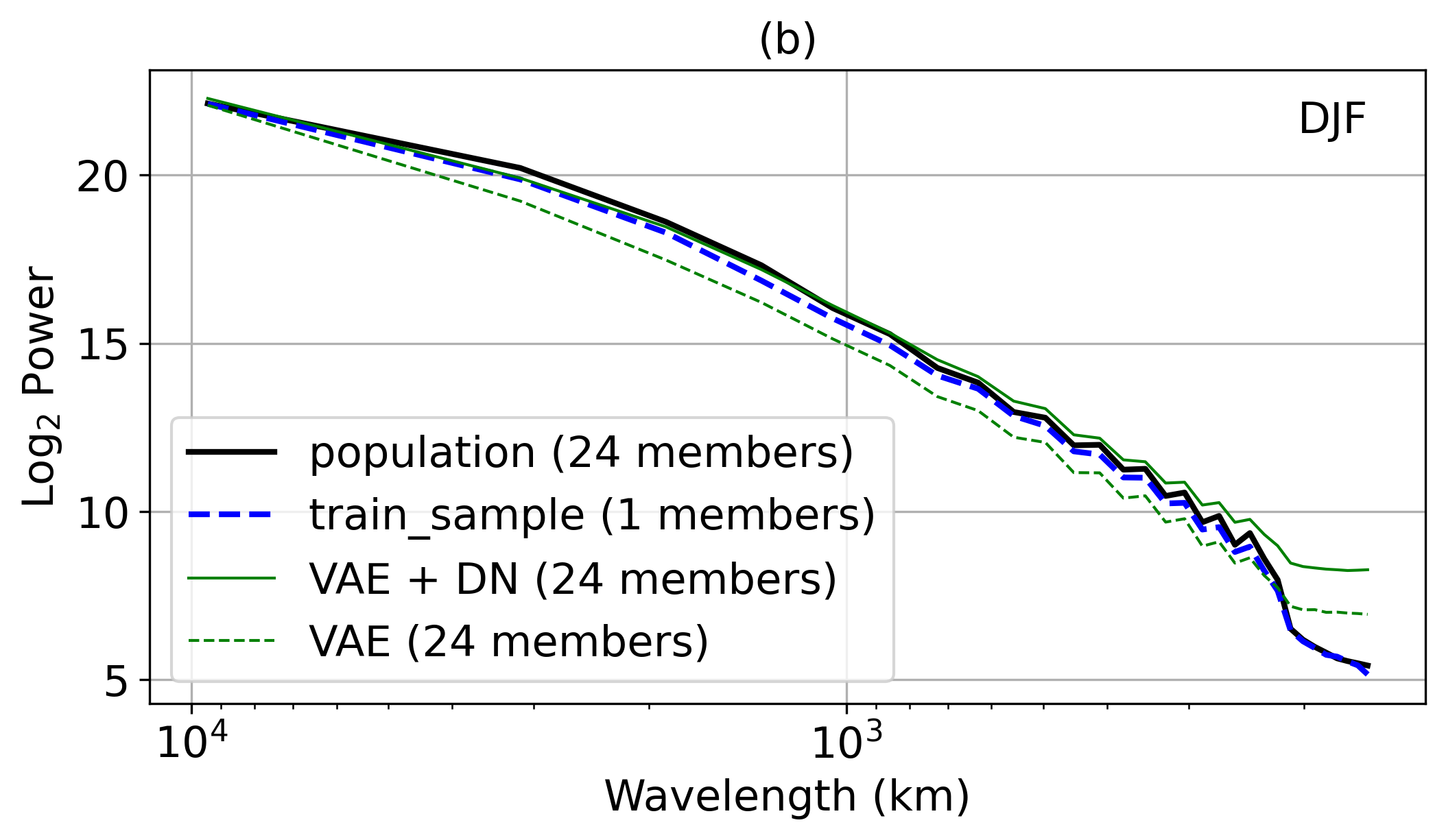}}
\end{minipage}\hfill
\begin{minipage}{0.48\textwidth}
    \centering
    \includegraphics[width=\linewidth]{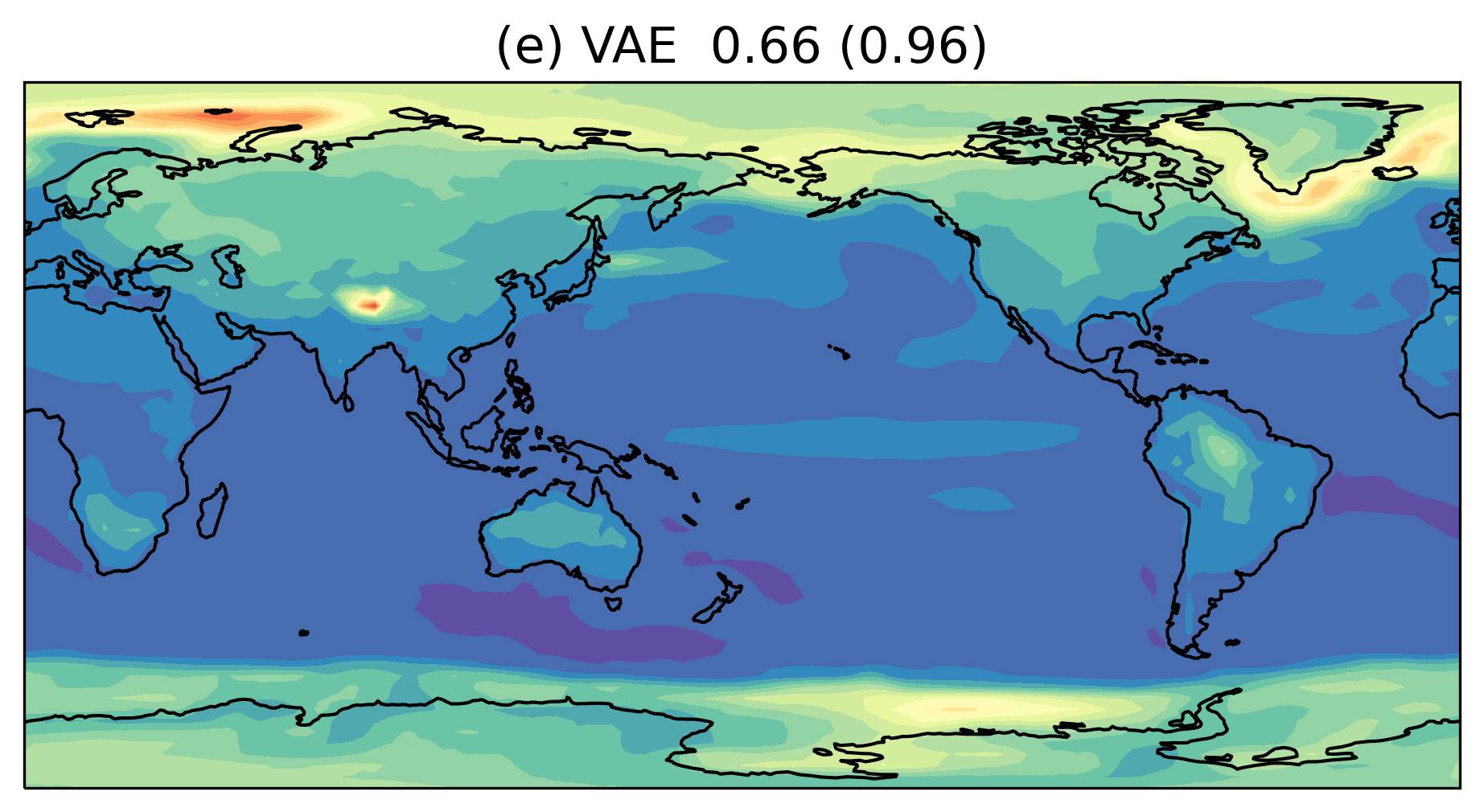}
\end{minipage}

\vspace{0.6em}

% ---------- Row 3 ----------
\begin{minipage}{0.48\textwidth}
    \centering
    \adjustbox{padding=0pt 0pt 0pt -10pt}{\includegraphics[width=\linewidth]{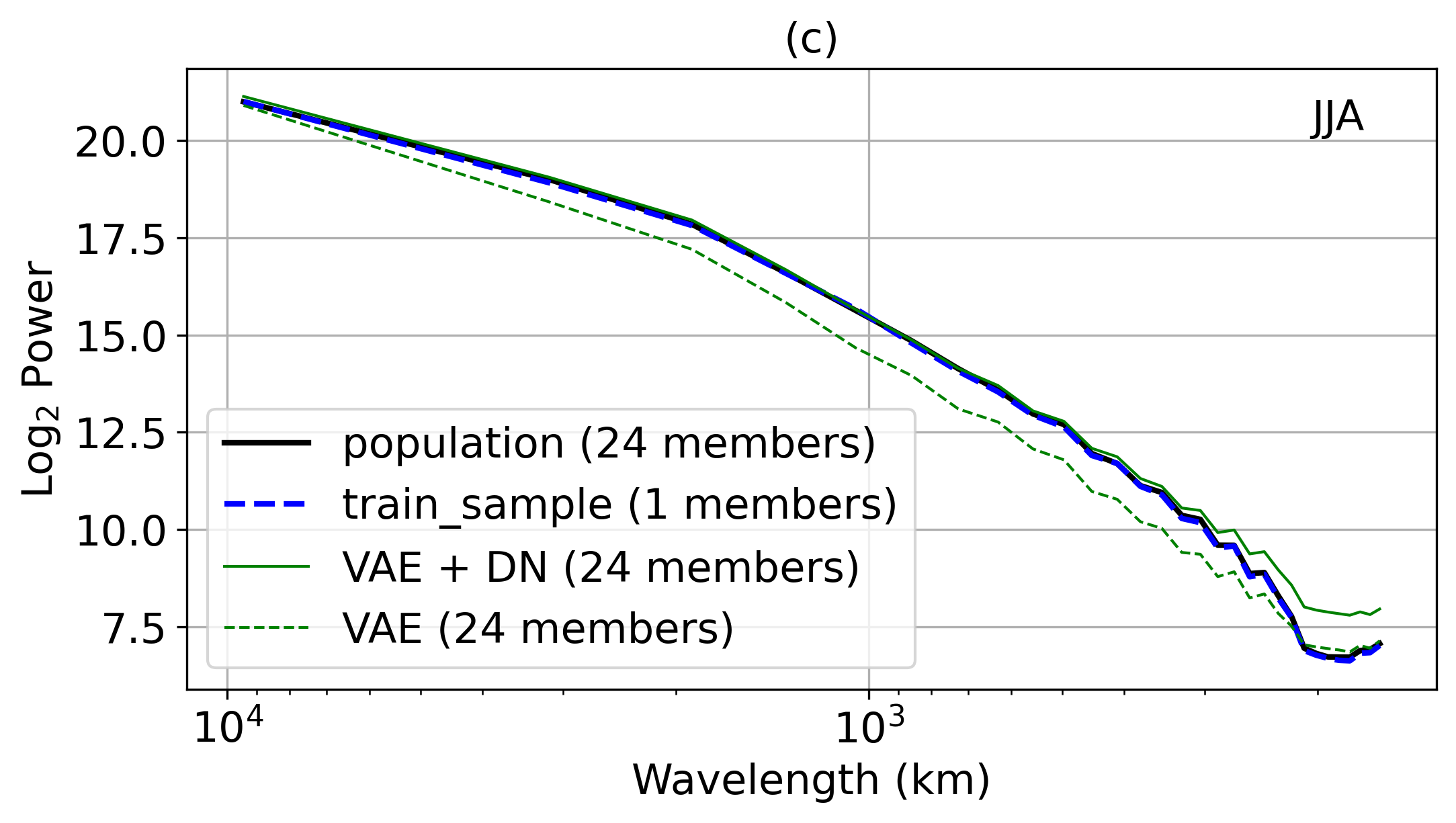}}
\end{minipage}\hfill
\begin{minipage}{0.48\textwidth}
    \centering
    \includegraphics[width=\linewidth]{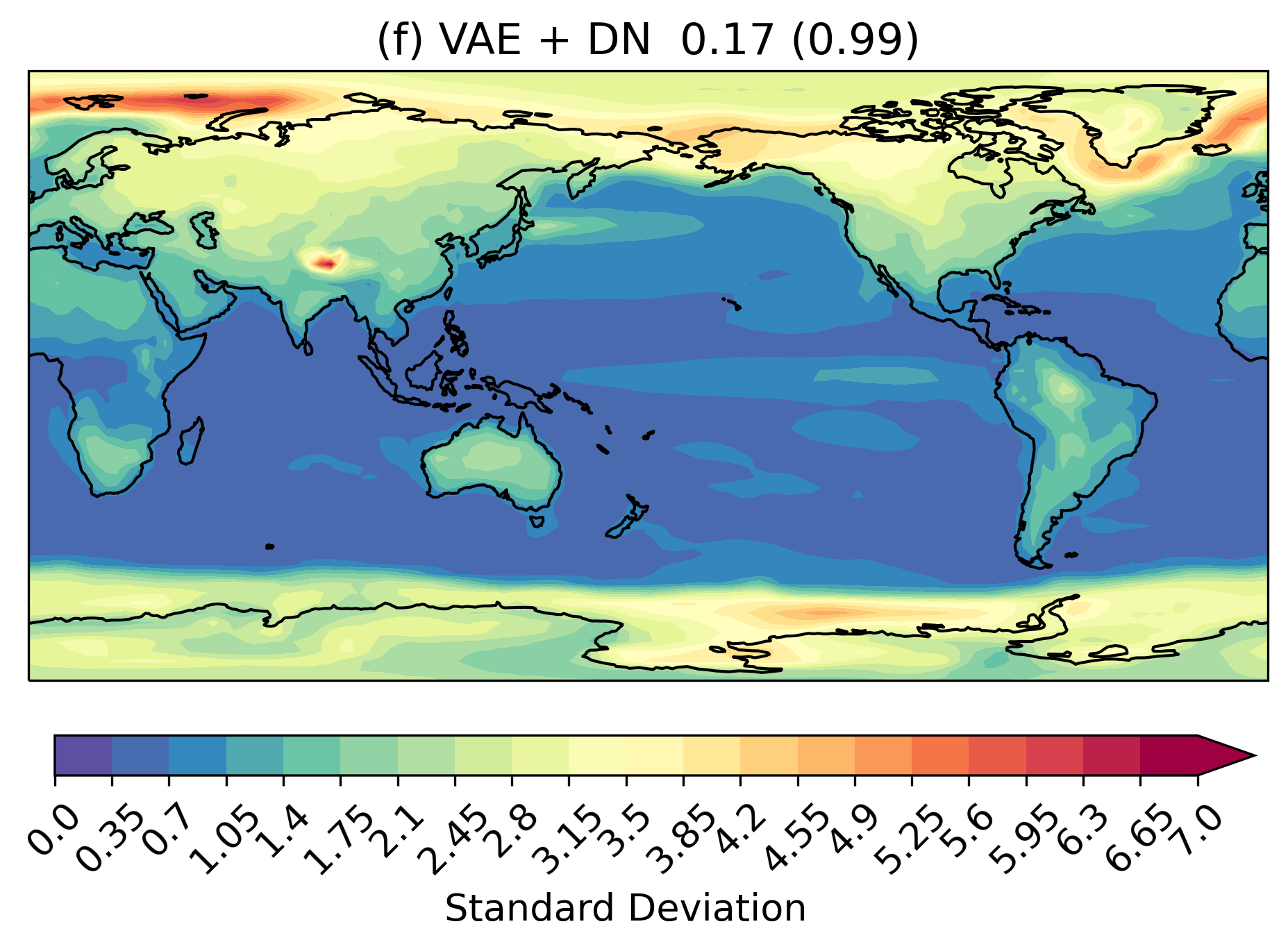}
\end{minipage}
\caption{ a) QQ plots of TAS anomalies relative to the seasonal climatology pooling all grid cells and times. b, c) Spectral power for DJF and JJA mean TAS anomalies averaged over years and realizations. d, e, f) standard deviation of TAS fields per grid cell across all times and realizations for different datasets. The numbers in the title of the panels are RMSE (pattern correlation) relative to the population on panel (d). All panels cover 1980-2020 period.} 
\label{fig:fig1}
\end{figure}

 To assess the impact of decoder noise, Fig. \ref{fig:fig1}a shows quantile-quantile (QQ) plots for the model output with decoder noise (VAE+DN) and without decoder noise (VAE) relative to the raw ensemble (population). The QQ plots are produced by pooling anomalies for all months and grid cells (i.e., after removing the sesonal cycle), with anomalies computed from monthly climatology over 1980-2020. We include the QQ plot for the anomalies in the training data for comparison. For a more complete assessment of variability, Figs. \ref{fig:fig1}b,c show the radially averaged power density (RAPSD) for the winter (DJF) and summer (JJA) seasons, while  Figs. \ref{fig:fig1}d-f show standard deviation maps. The maps show standard deviation at each location based on the pooled anomalies, while RAPSD is computed for each seasonal mean anomaly map by radially averaging its 2D Fourier spectra around the origin of the frequency domain \cite{nbsgf17}. 

For VAE, the model underestimates the variability of the population, as indicated by the relatively shallow slope of the QQ curve (Fig. \ref{fig:fig1}a). This muted variability is most pronounced on land and in the extratropical oceans (Figs. \ref{fig:fig1}d,e), but it is also underestimated in the tropical Pacific, implying a relatively weak representation of ENSO in the VAE ensemble. In contrast, the addition of decoder noise in VAE+DN helps counteract this reduced variability, leading to clear improvements globally (Figs. \ref{fig:fig1}d,e,f) and across both the lower and upper percentiles of the anomaly distribution (Fig. \ref{fig:fig1}a). The power spectra curves show (Fig. \ref{fig:fig1}b,c) an overall improvement in spatial variability for VAE+DN compared to VAE (which is mostly biased low) or the training sample, most notably during winter, indicating the benefits of decoder noise. However, for finer spatial scales (below $\sim200$ kms), VAE+DN overestimates the population variability. While the coarse  $64\times128$ grid size of TAS fields makes the estimation of power over fine resolution noisy, the error could also reflect the coarse resolution of CanESM5 and the limited ability of the raw ensemble to capture true fine-scale variation (section \ref{subsec:data_n_prepro}). Overall, our results show that the addition of decoder noise consistently enhances model performance, and therefore we adopt the VAE+DN configuration as our default going forward, unless otherwise noted. We expect more expressive architectures and higher spatial resolution in the data to improve the biases in the original VAE decoder. However, the VAE decoder can be improved insofar as the predictability limit of the system has not been reached \cite{doublepenalty}.

\section{Results and discussions}
\label{sec:results_n_discussions}

\subsection{Regional statistics}
\label{subsec:regions}

% The main purpose of boosting the ensemble size of climate simulations is to have better assessment of distributional properties of the climate system and proper representation of extremes which is normally not possible using a single realization (or small ensemble size). This is specially important for populated regions where extreme temperatures could put human life in great risk. As a first test, we compare the boosted ensemble using the VAE+DN to the full CanESM5 ensemble (population) over the period of 1980-2020 based on anomalies relative to the monthly climatology over four major cities - Vancouver, London, Hong Kong, and Los Angeles using QQ plots and probability density functions (PDF). Figure 2 panels a,c,e and g show that in all cases, the VAE+DN has distributional properties in close agreement to the population as indicated by the the QQ plots being close to the 1:1 line. Interestingly, in almost every case, the boosted ensemble is in closer agreement to the population compared to the limited training sample. One exception is Vancouver where the higher quantiles are larger for the VAE+DN indicating possible overestimation/overdispersion at this location. We will use PDF plots to take a closer look at this behavior.

\begin{figure}
\centering

% ---------- Row 1 ----------
\begin{minipage}{0.24\textwidth}
    \centering
    \includegraphics[width=\linewidth]{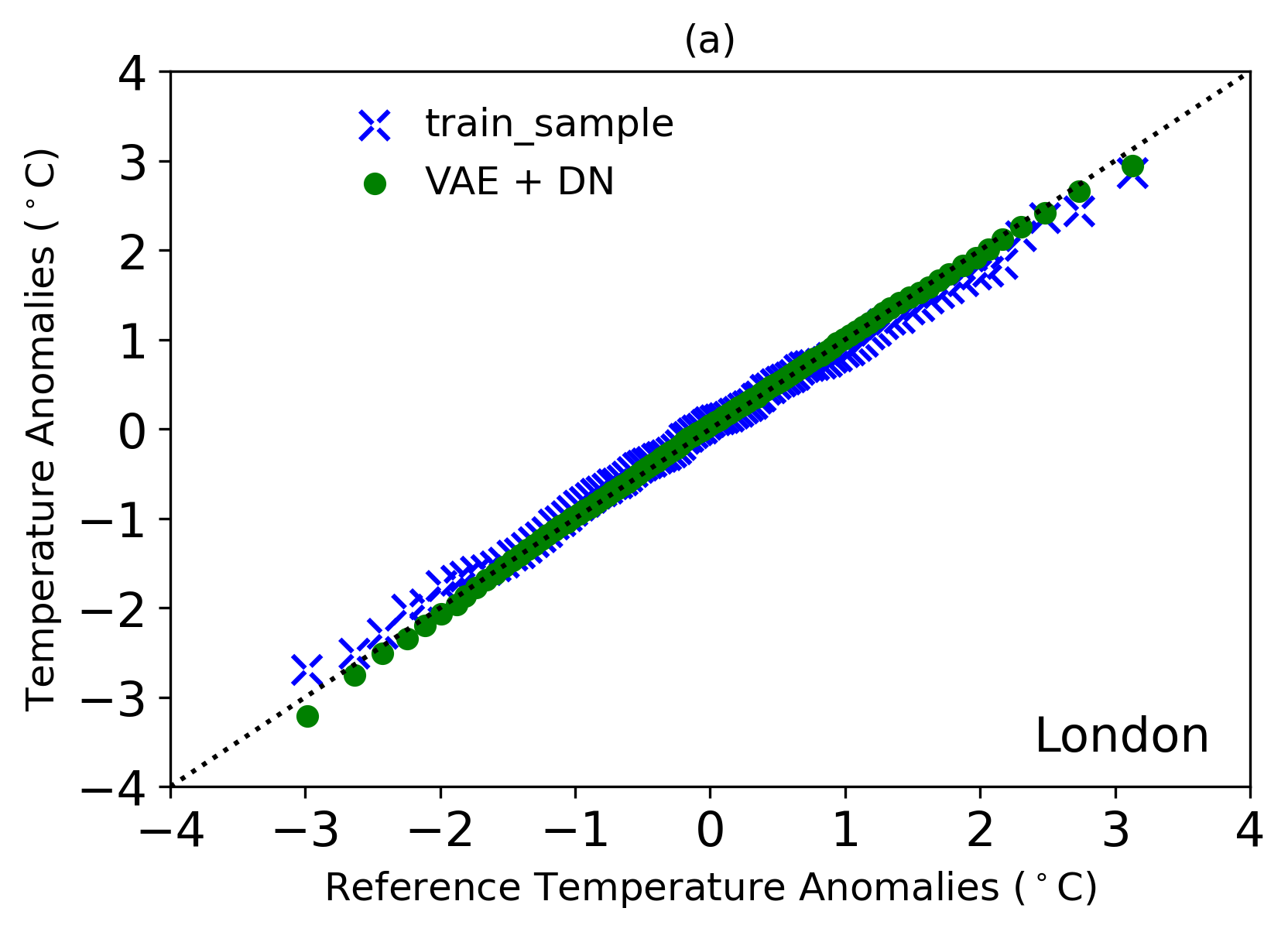}
\end{minipage}\hfill
\begin{minipage}{0.24\textwidth}
    \centering
    \includegraphics[width=\linewidth]{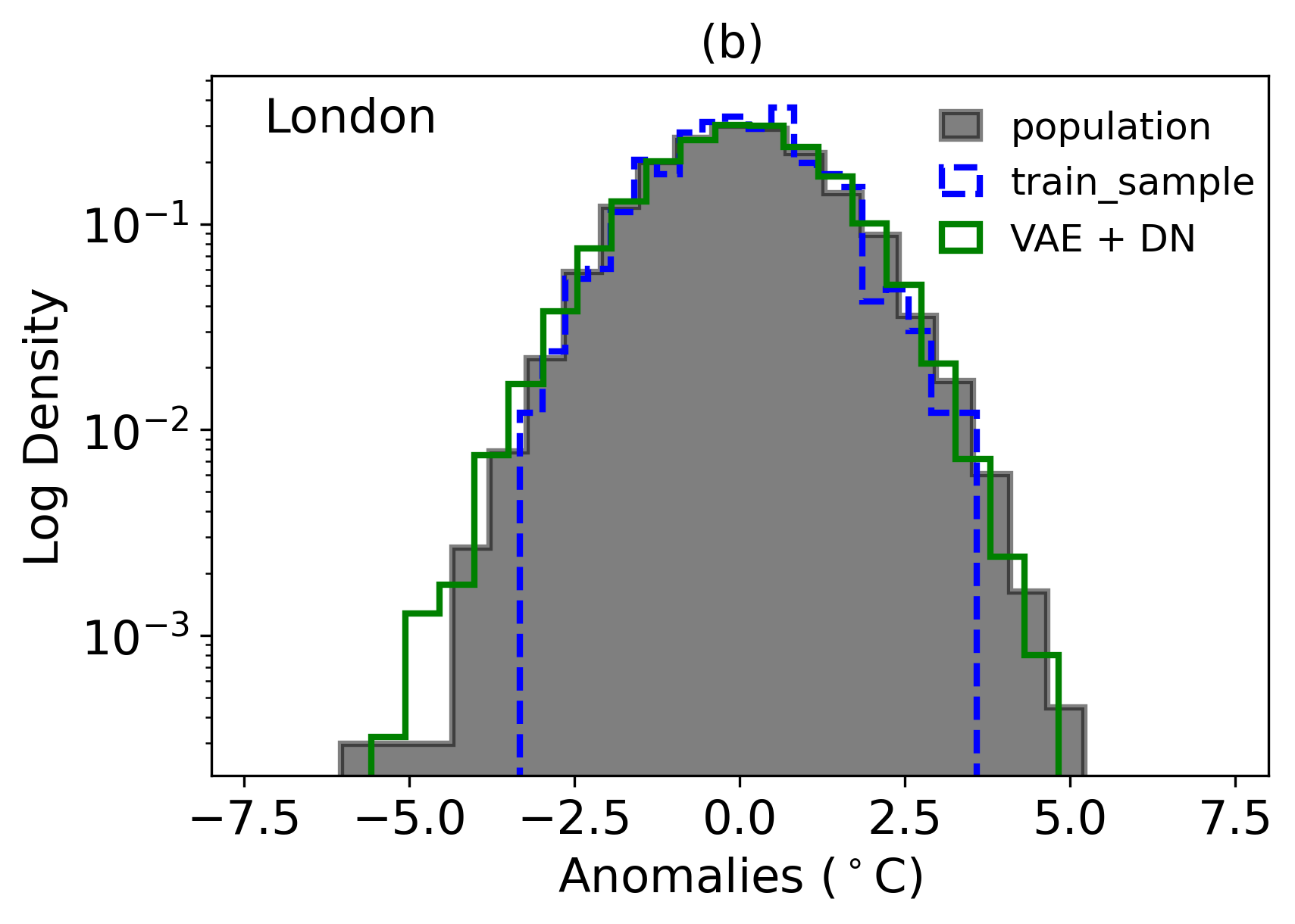}
\end{minipage}\hfill
\begin{minipage}{0.24\textwidth}
    \centering
    \includegraphics[width=\linewidth]{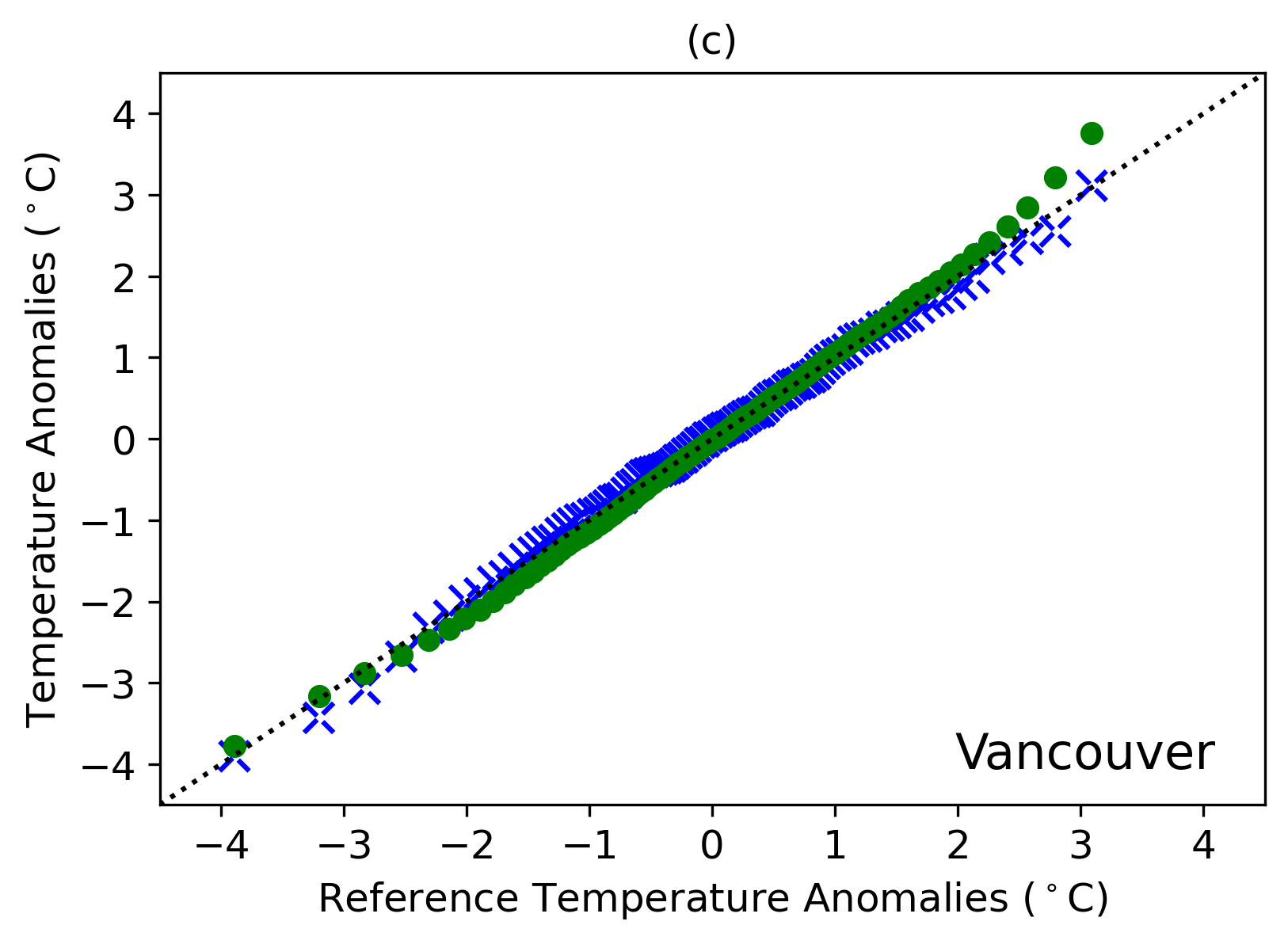}
\end{minipage}\hfill
\begin{minipage}{0.24\textwidth}
    \centering
    \includegraphics[width=\linewidth]{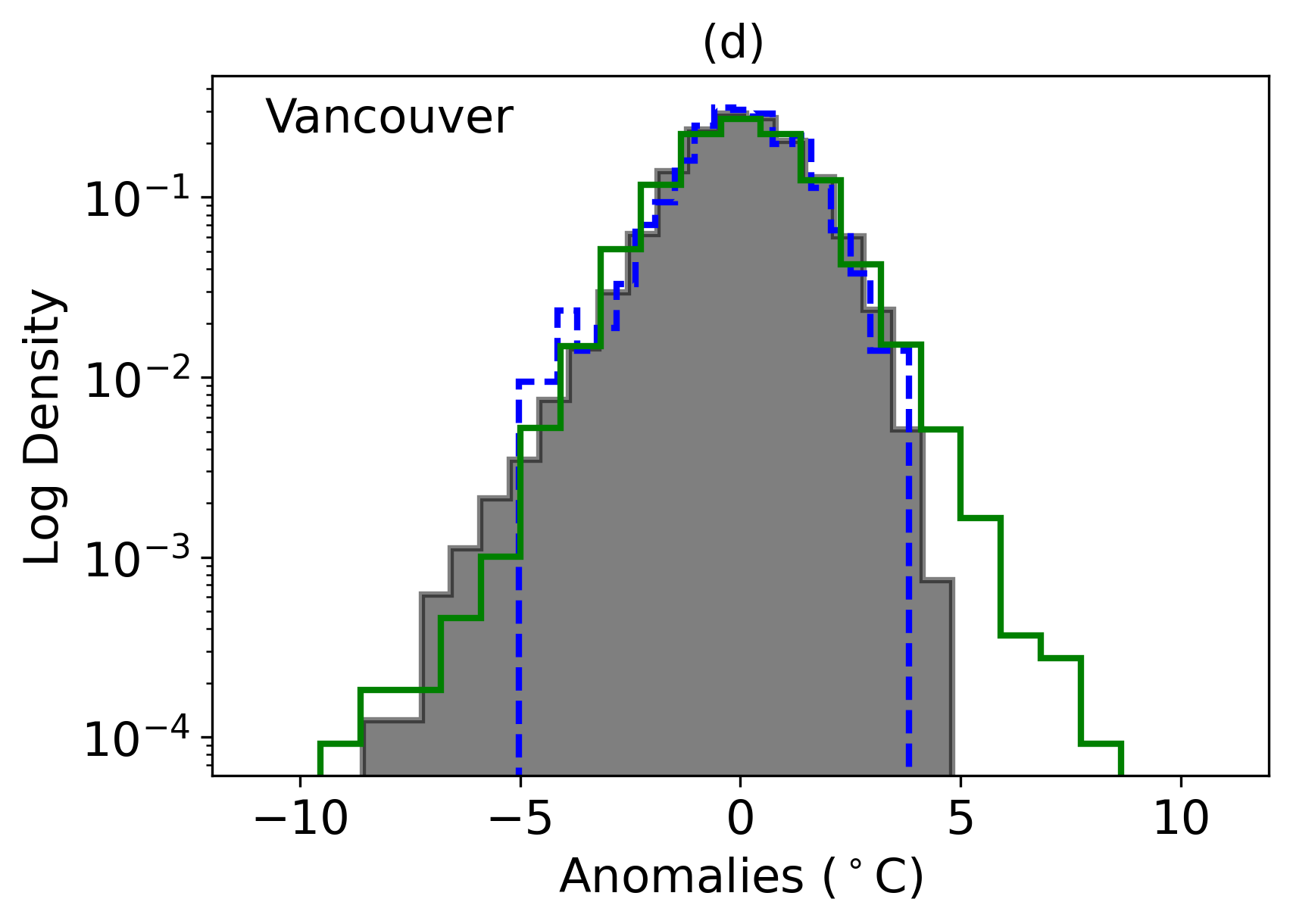}
\end{minipage}

% \vspace{0.5em}

% ---------- Row 2 ----------
\begin{minipage}{0.24\textwidth}
    \centering
    \includegraphics[width=\linewidth]{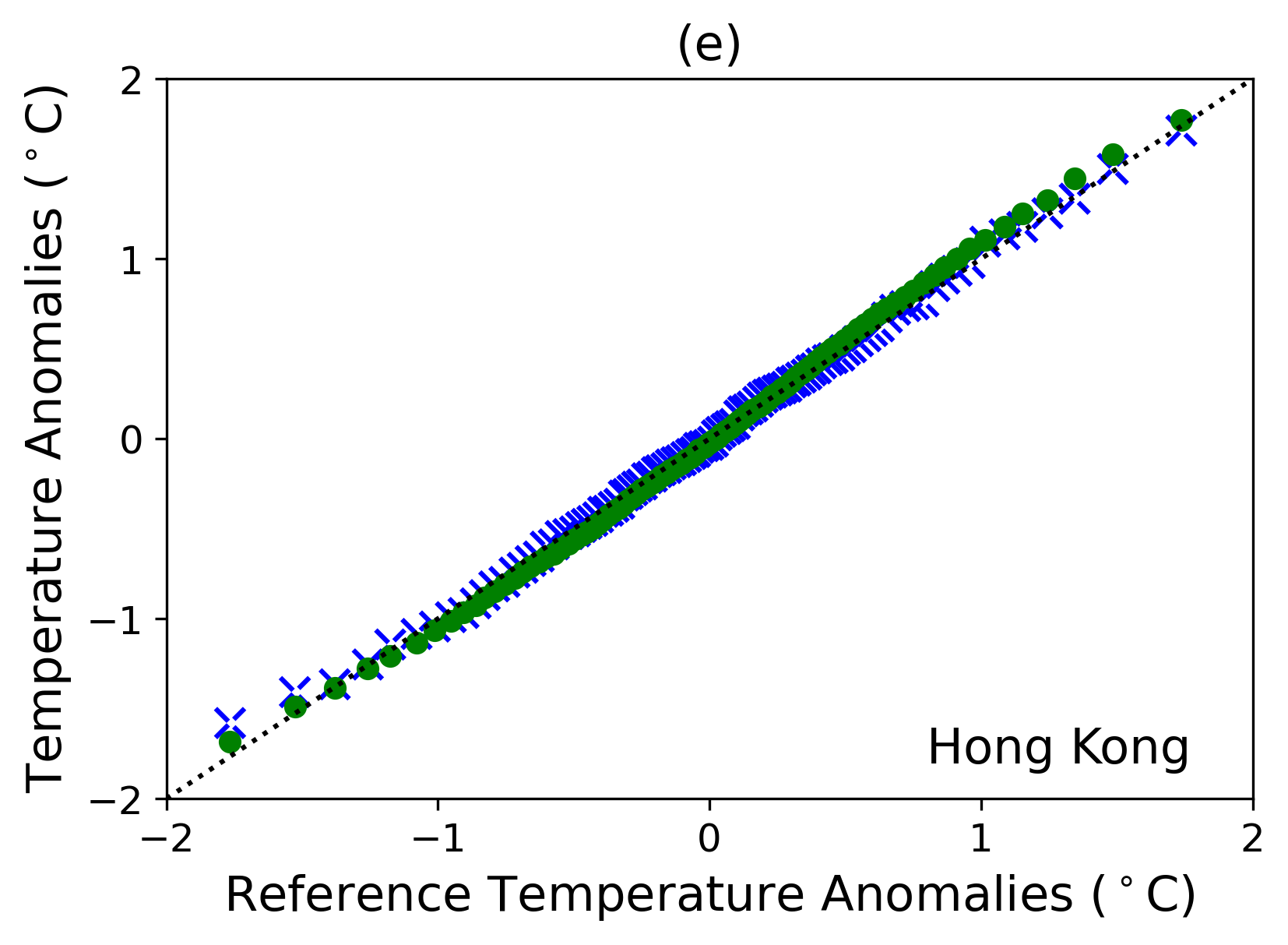}
\end{minipage}\hfill
\begin{minipage}{0.24\textwidth}
    \centering
    \includegraphics[width=\linewidth, trim=0 0 125 0, clip]{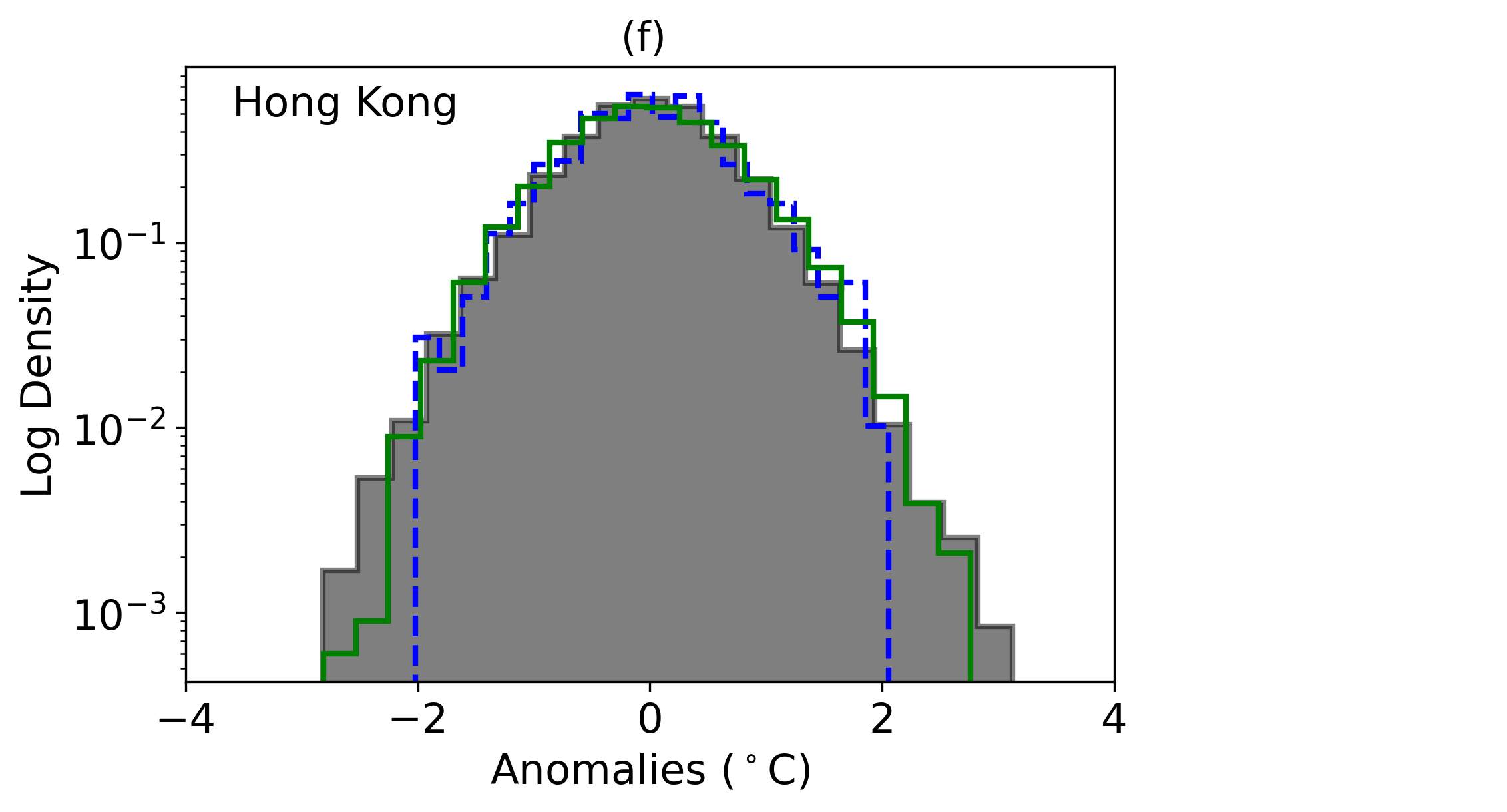}
\end{minipage}\hfill
\begin{minipage}{0.24\textwidth}
    \centering
    \includegraphics[width=\linewidth]{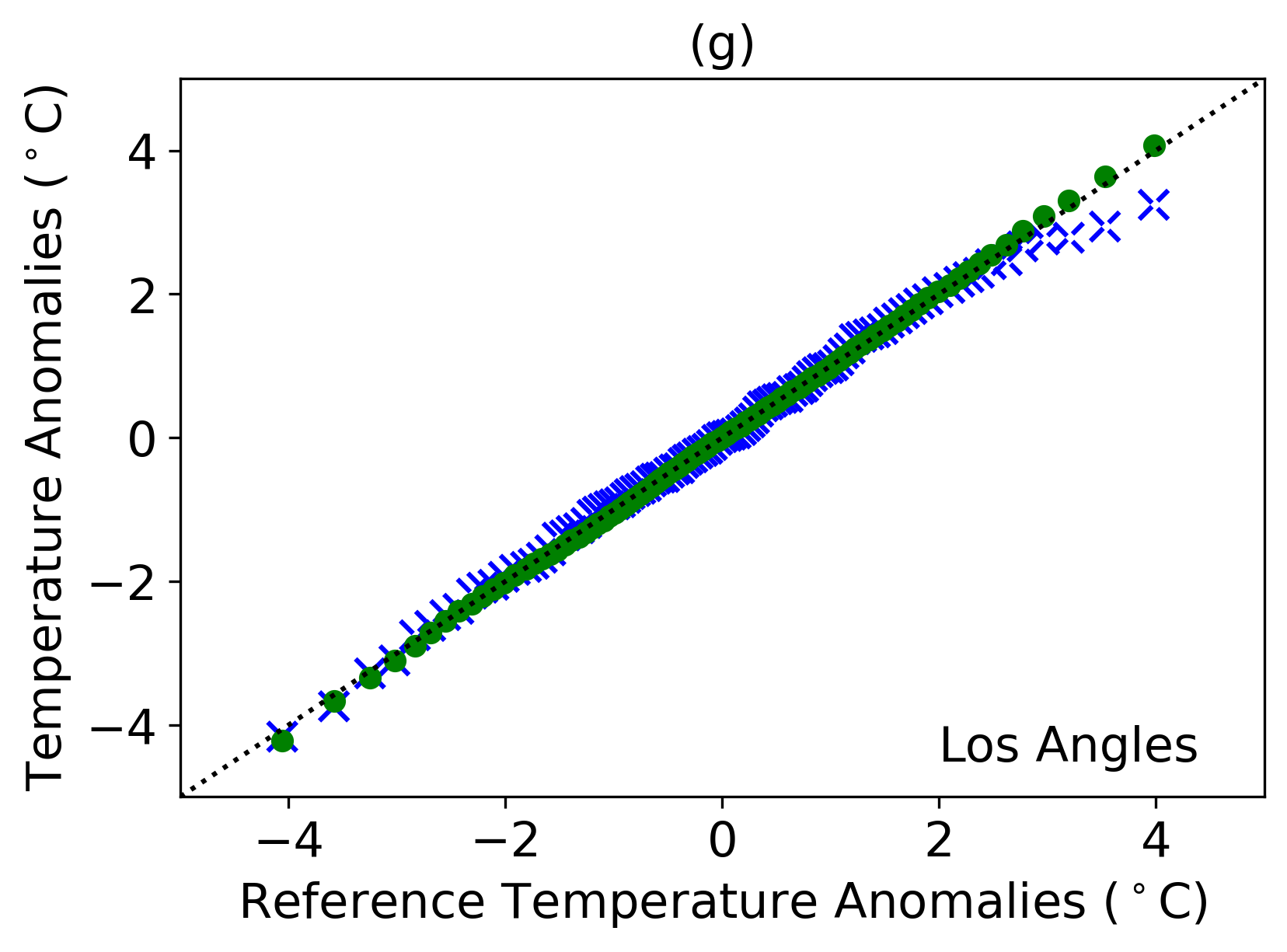}
\end{minipage}\hfill
\begin{minipage}{0.24\textwidth}
    \centering
    \includegraphics[width=\linewidth, trim=0 0 125 0, clip]{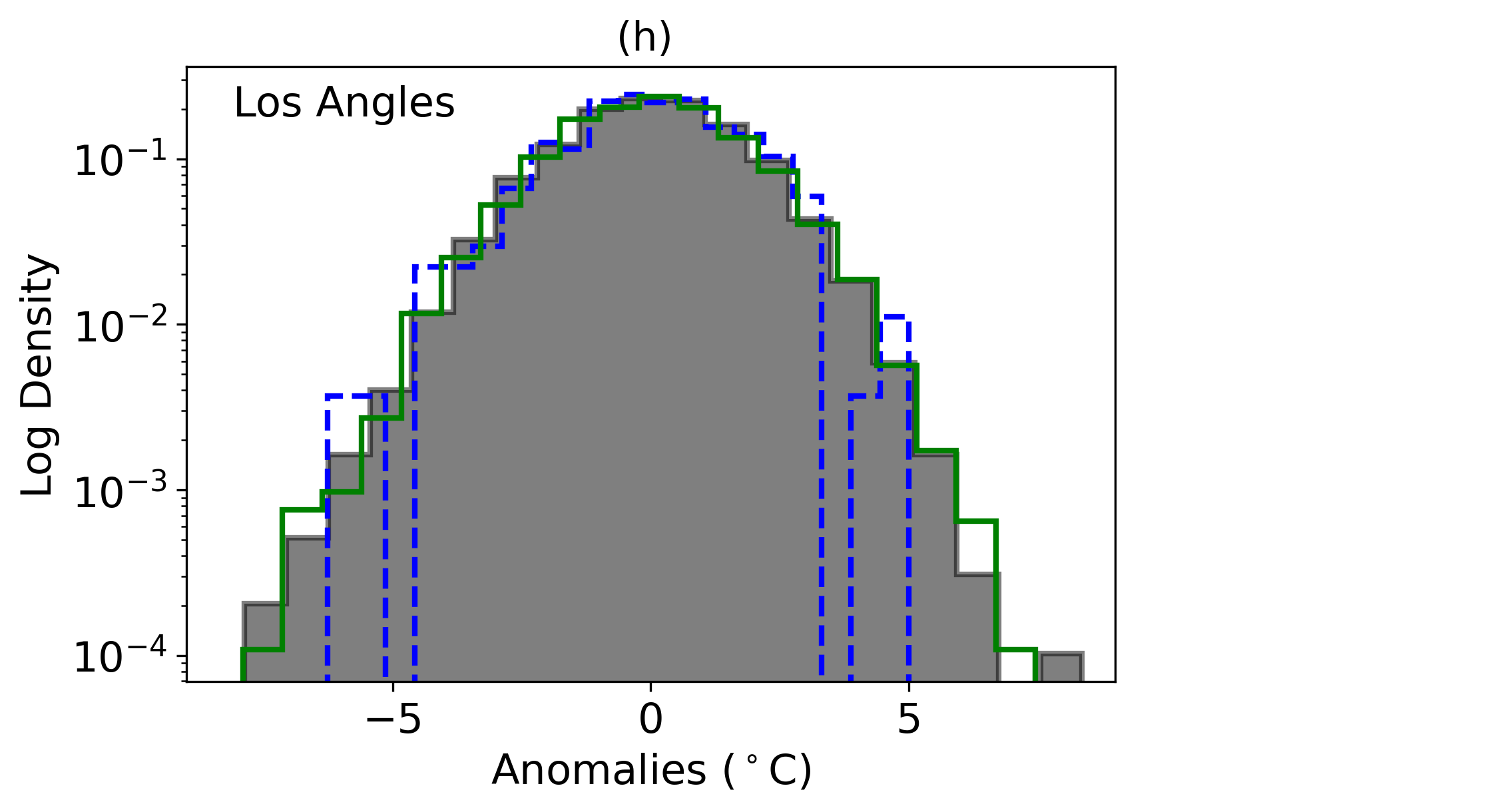}
\end{minipage}

\caption{ a, c, e, g) QQ plots of TAS anomalies relative to the seasonal climatology at different cities. b, d, f, h) PDF histograms on log scale for the same cities.} 
\label{fig:fig2}
\end{figure}

The main purpose of boosting climate simulation ensembles is to accurately assess statistical properties of climate events, including extremes, which is not possible using a single realization or small ensembles. This is especially important for populated regions, where extreme events can lead to widespread disruptions. As an initial evaluation, we compare the VAE+DN boosted ensemble with the full CanESM5 ensemble (population) over the 1980-2020 period for model grid cells covering Vancouver, London, Hong Kong and Los Angeles, through QQ plots and probability density functions (PDFs) of anomalies from monthly climatology (Fig. \ref{fig:fig2}).

The QQ plots (Fig. \ref{fig:fig2}a,c,e,g) show that the VAE+DN model reproduces the percentiles of the anomaly distribution in the population, generally improving over the training sample for both the lower and upper percentiles. One exception is Vancouver, where the VAE+DN model overestimates the upper percentiles for reasons we examine below. To asses the model's ability to capture the distribution tails and extremes in the population, Fig. \ref{fig:fig2}b,d,f,g shows the log PDFs for all locations. Overall, the boosted ensemble successfully captures events in the population that are absent from the training data, indicating the VAE+DN model's ability to generalize for out-of-training data. In the case of Vancouver, the model clearly represents events in the distribution tail that are not present in the population. This may be in part due to (1) limited training data being non-representative of possible asymmetric behavior (as seen in Fig. 2d) (2) the decoder tendency to produce symmetric distributions or (3) because the model generates relatively strong warming possibly tied to a learned relationship in the data. As we will see in the next section, Vancouver is within the ENSO teleconnection region known for El Ni\~no-induced warming. Regarding (2), we note that the boosted ensemble is nevertheless able to capture non-Gaussian behavior at these locations as shown with some detail in the supplements (Fig. S2).

\begin{figure}
\centering

\begin{minipage}{0.32\textwidth}
    \centering
    \includegraphics[width=\linewidth]{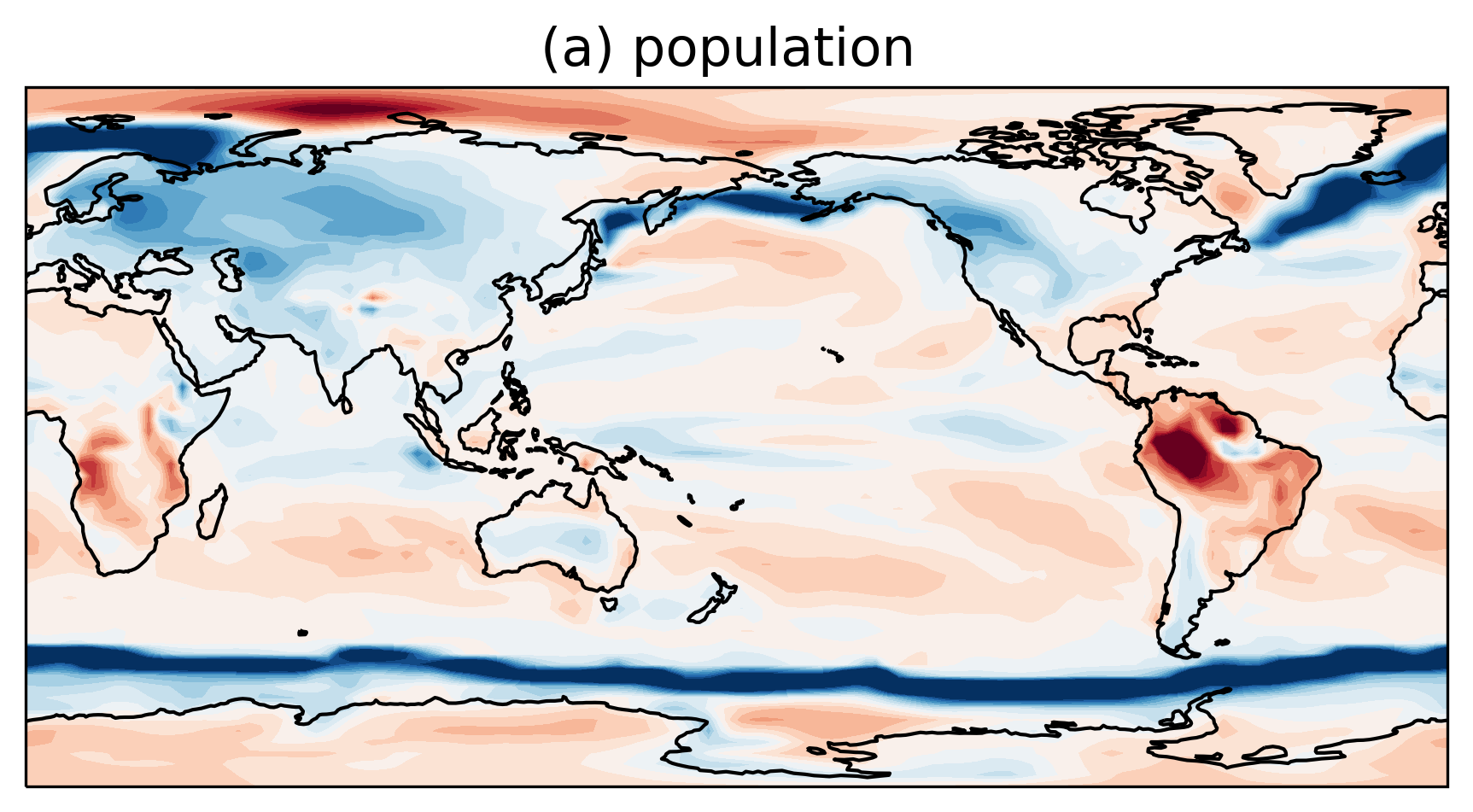}
\end{minipage}
\hfill
\begin{minipage}{0.32\textwidth}
    \centering
    \includegraphics[width=\linewidth]{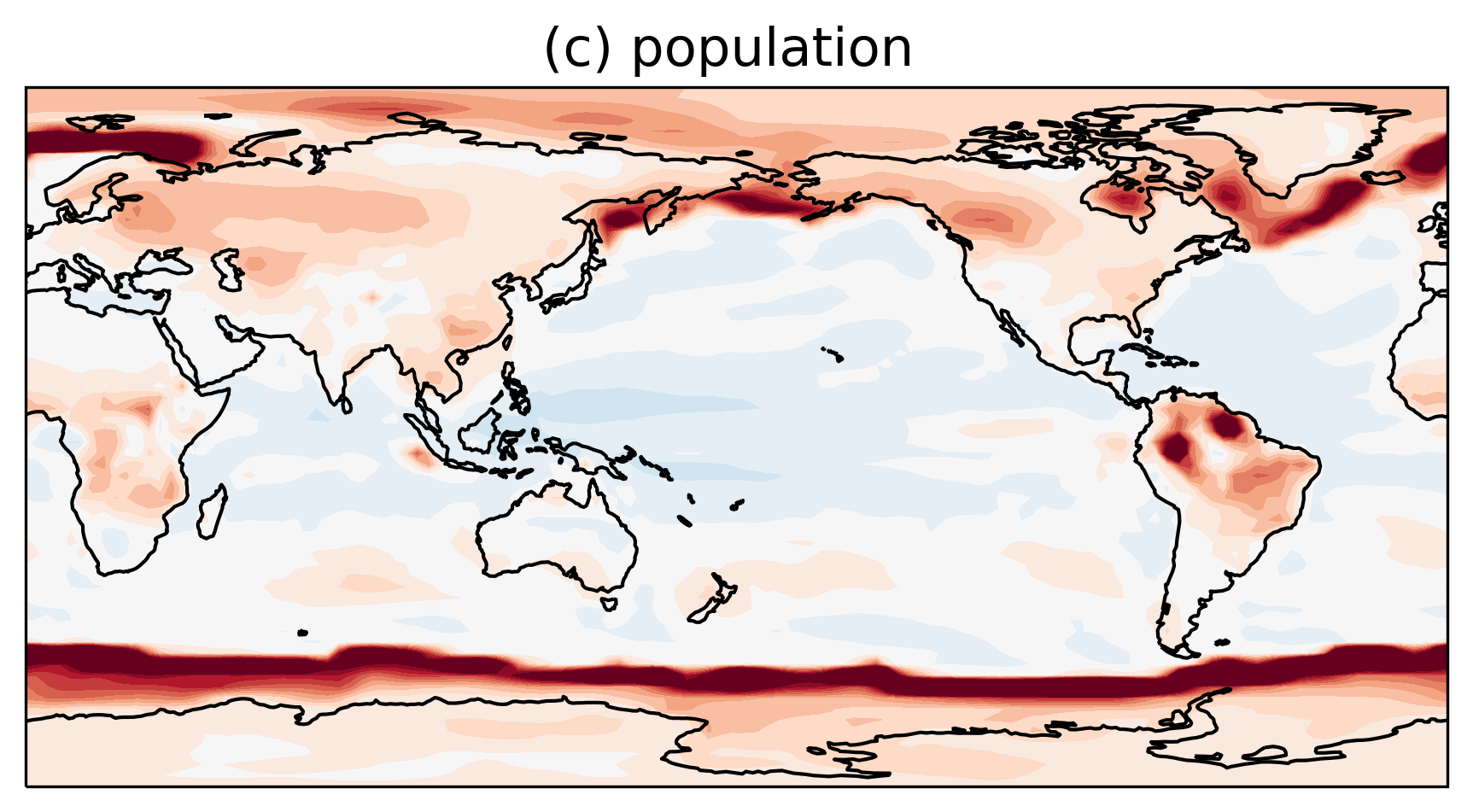}
\end{minipage}
\hfill
\begin{minipage}{0.32\textwidth}
    \centering
    \includegraphics[width=\linewidth]{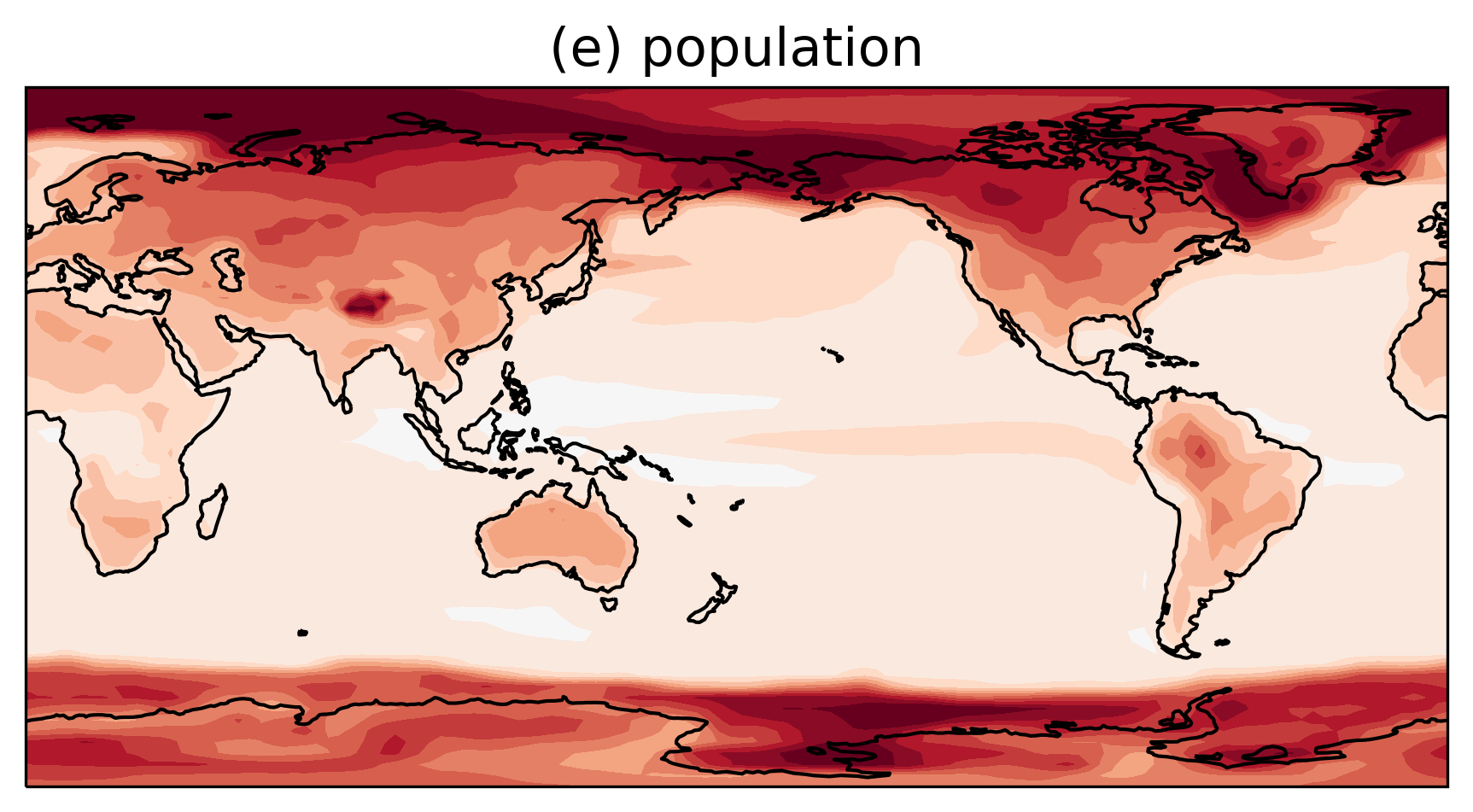}
\end{minipage}

\begin{minipage}{0.32\textwidth}
    \centering
    \includegraphics[width=\linewidth]{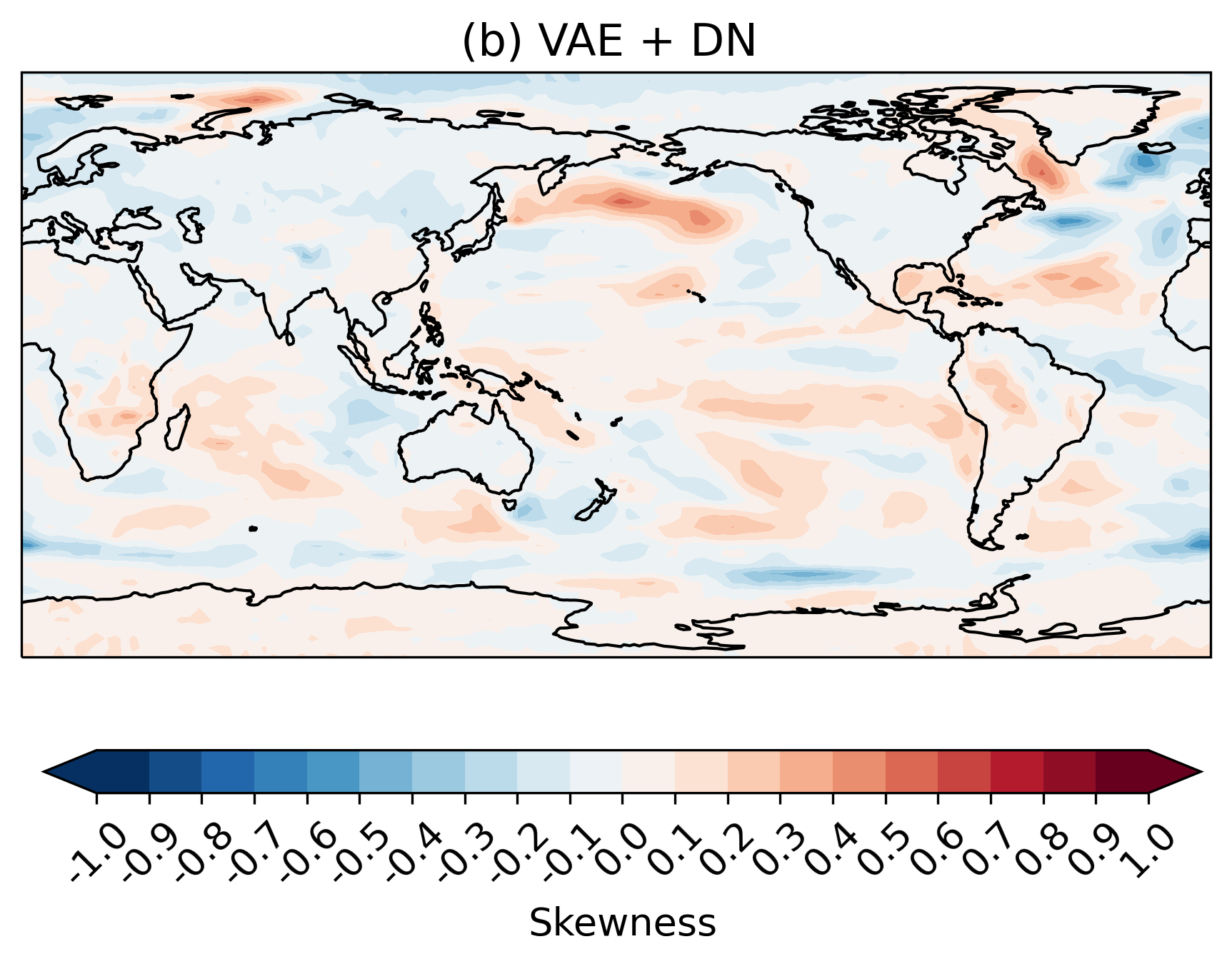}
\end{minipage}
\hfill
\begin{minipage}{0.32\textwidth}
    \centering
    \includegraphics[width=\linewidth]{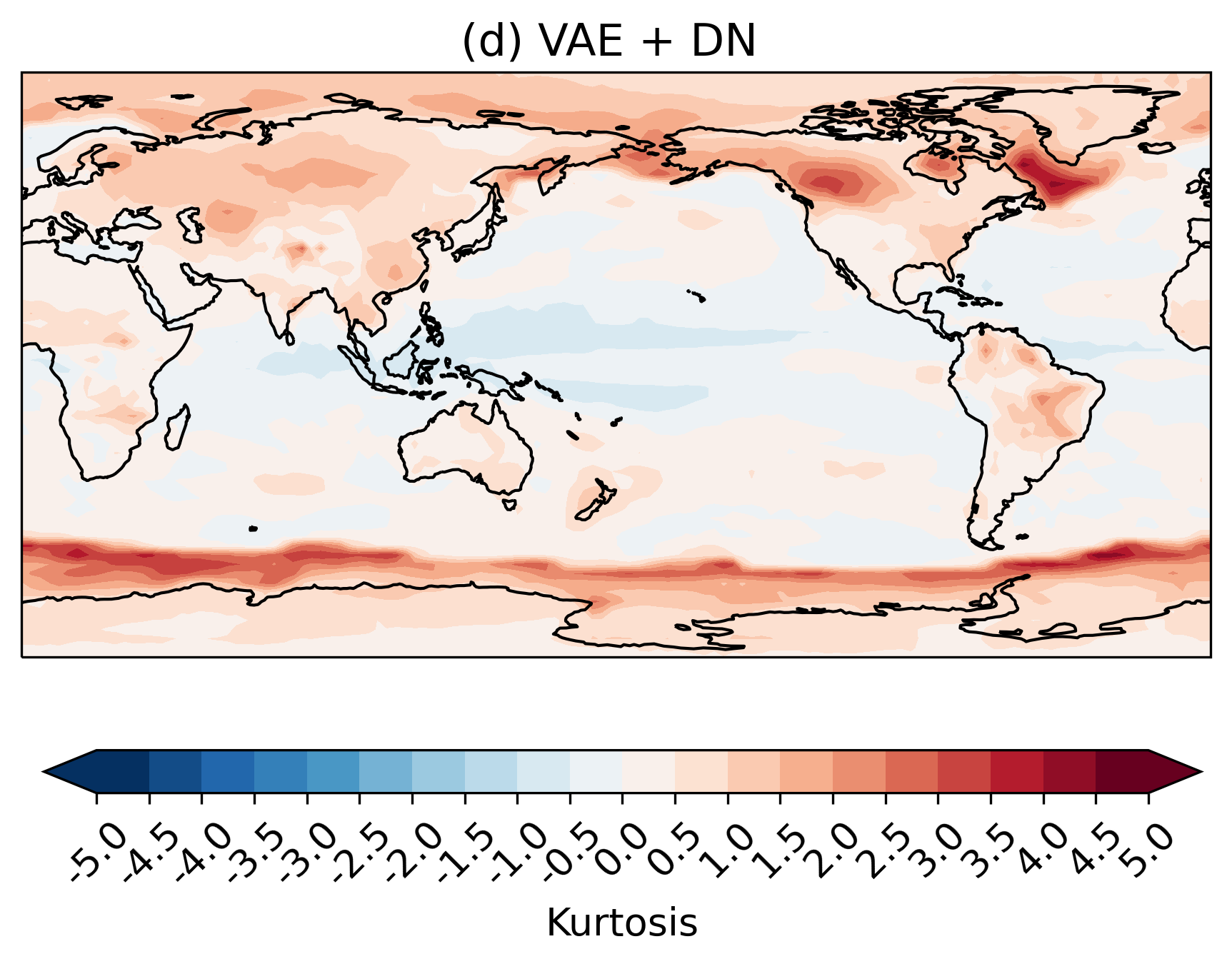}
\end{minipage}
\hfill
\begin{minipage}{0.32\textwidth}
    \centering
    \includegraphics[width=\linewidth]{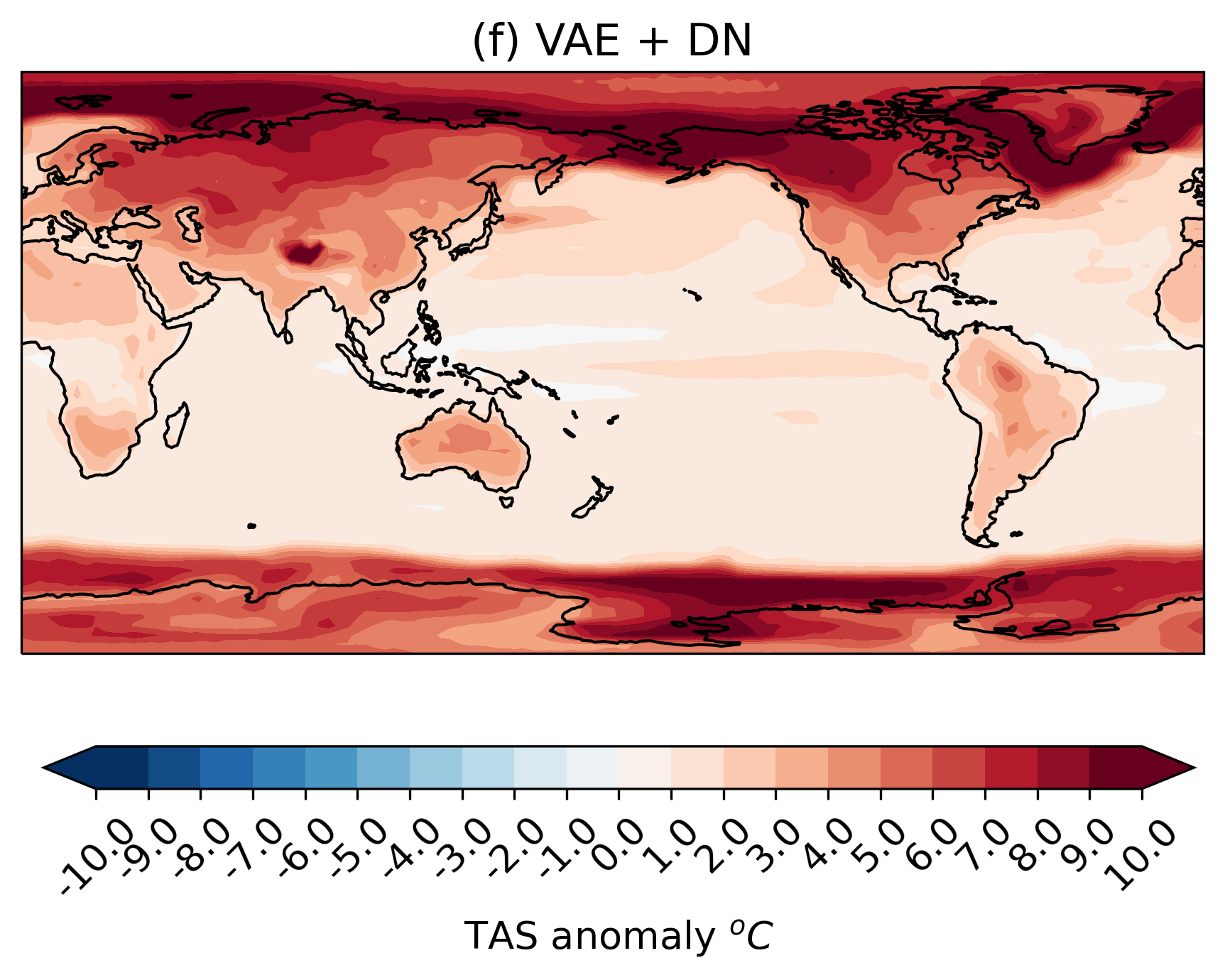}
\end{minipage}

\caption{ a, b) Maps of skewness (c), (d) kurtosis and (e), (f) \%0.99 percentile of TAS anomalies for population (top row) and VAE+DN (bottom row)  } 
\label{fig:fig3}
\end{figure}

To further assess the ability of the boosted ensemble to represent the TAS distribution in the population, particularly potential asymmetries and non-Gaussain behavior, Figs. \ref{fig:fig3}a-d show measures of the third and fourth moments of the TAS distribution in the form of skewness and excess kurtosis. Figures \ref{fig:fig3}a,c show that the population is characterized by higher latitudes with non-symmetric distributions and heavy tails, coincident with regions of strong variability (Fig. \ref{fig:fig1}d). The VAE+DN boosted ensemble captures these global patterns but with underestimated magnitudes (Figs. \ref{fig:fig3}b,d). This underestimation is not uncommon, even for more expressive architectures and larger training samples \cite{wang2025gen2generativepredictioncorrectionframework}, although our limited training data may play a role. We examine whether our approach to estimate the noise in the decoder as a Gaussian with a constant covariance matrix independent of the generated output (Eq. \ref{eq:cVAE_updated}), impacts skewness and kurtosis (Fig. S3 of the supplements). The results confirm that the asymmetric behavior of the distribution is better represented without decoder noise, while kurtosis (tail behavior) is largely underestimated possibly due to the loss of variability associated with the smoothed output of the decoder ${\mu}_{NN_{\theta}}(z,c)$ (section \ref{subsec:inference}). We note that, although the VAE parameterizes its latent space as Gaussian, this does not preclude the model from capturing non‑Gaussian behavior in its outputs. While the above results clearly prove this point, we emphasize that the Gaussian formulation of the latent space does not translate to the output since the decoder transformation is non-linear. All of the above suggests that there is a trade-off in using the decoder noise under the simple formulation of Eq. \ref{eq:sigma_train_x}. This becomes more important for variables such as precipitation with stronger non-Gaussian behavior where more expressive models, or better formulation of decoder noise might be required. Given that all other relevant metrics examined here are largely improved and in close agreement with the population in the VAE+DN model (section \ref{subsec:DN}), especially with respect to underdispersivity, and given the  importance of representing and quantifying extremes, we encourage the use of decoder noise despite its simplicity and its impact on the skewness of the distribution.

Further evaluation of extremes reinforce this point. The composite of the 99-percentile TAS anomaly maps, computed separately for the population and boosted ensemble (Figure 3c,d), show close agreement between the two (0.97 pattern correlation and 0.85 $^\circ C$ RMSE). VAE+DN reproduces the large TAS anomalies of the higher latitudes, although it has warm biases in a few regions including the Labrador Sea and Southern Ocean (Fig. S4). Comparing with the skewness maps (Fig. \ref{fig:fig3}), these are regions where the negative asymmetry in the population is underestimated in the VAE+DN (similar to the behavior for Vancouver in Fig. \ref{fig:fig2}d) giving rise to the overestimation of the 99-percentile. Regardless, there is clear and meaningful agreement in both global patterns and magnitude. 

In summary, these comparisons show that the simple VAE+DN model is able to generate events outside the training sample, even in regions where non-Gaussian behavior exists. This indicates that the model learns the underlying distribution of the data and generates realizations that are representative of the population. As the skewness maps show, the largest discrepancies are in capturing asymmetries in distributions, leading to overestimation of extremes in some locations. While these results are satisfactory, we anticipate that a more robust treatment of decoder noise, whether through an improved noise formulation or more expressive VAE architectures leading to less structured residuals (i.e., a more accurate decoder mean), will further enhance performance. In this section, we evaluated regional statistics. Next, we evaluate whether the model captures global dependencies and physically-meaningful patterns and teleconnections resulting from dominant modes of climate variability.

\subsection{Global structure and teleconnections}
\label{subsec:enso}

VAE models (more generally models with MSE loss function) are trained to maximize the log likelihood of the marginal (here conditional) distributions, i.e, reference value at each grid cell. An important test to assess whether the generated fields do not result from hallucination or overfitting to the marginals is to check if they represent physically meaningful patterns and cross-location dependencies. El Niño Southern Ocean oscillation (ENSO) is the dominant mode of climate variability on seasonal to interannual time scales, driving  regional weather and climate conditions, including extremes, through well-known teleconnections. A widely used indicator of ENSO is the Niño3.4 index, which we compute as the monthly TAS anomalies averaged over the Niño 3.4 region (5S-5N, 170W-120W). We start by assessing the statistical properties of the generated Niño 3.4 index.

\begin{figure}
% \vspace*{-3.cm} 
\begin{minipage}{0.32\textwidth}
    \centering
    \includegraphics[width=\linewidth]{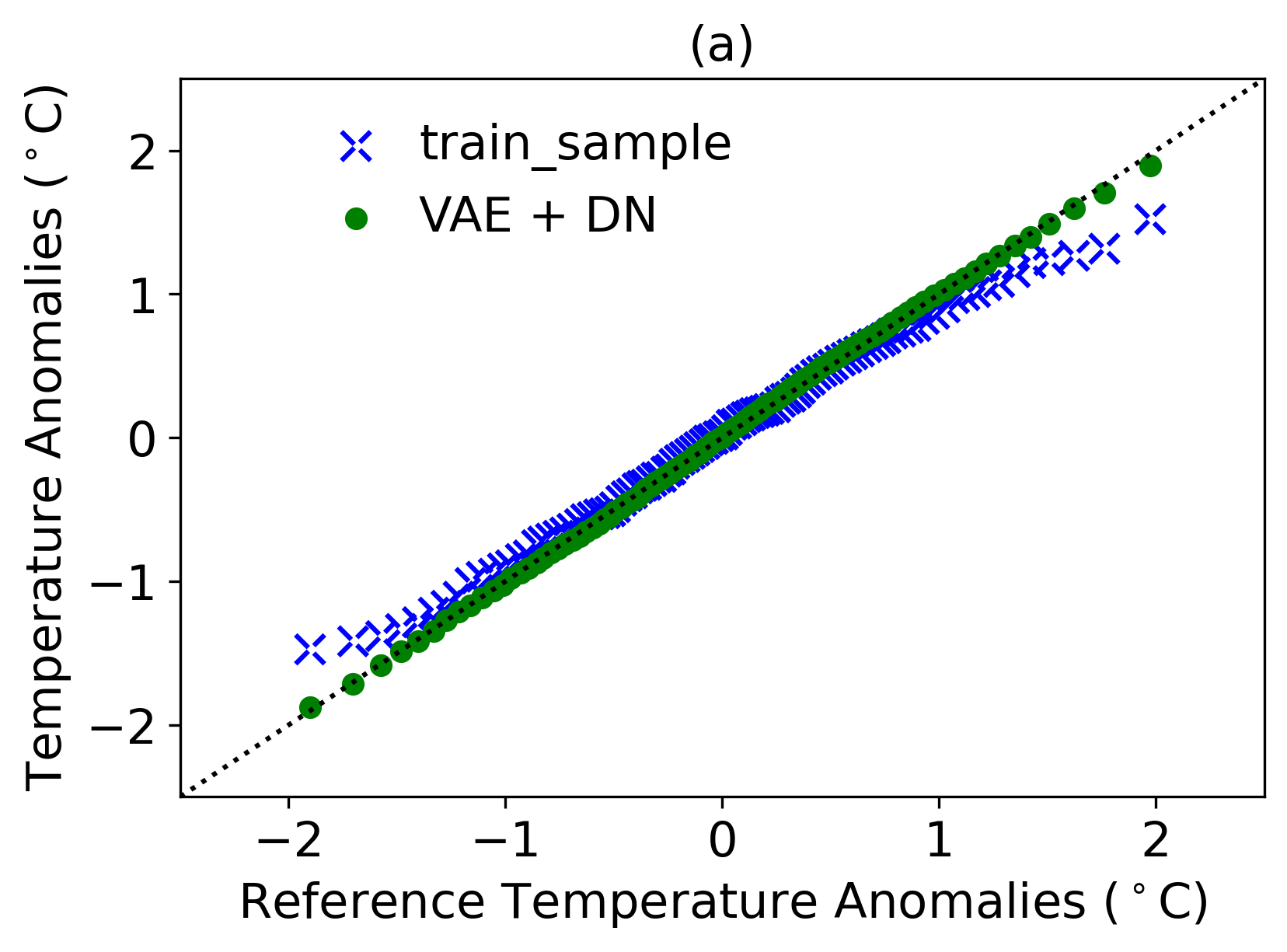}
\end{minipage}
\hfill
\begin{minipage}{0.32\textwidth}
    \centering
    \includegraphics[width=\linewidth]{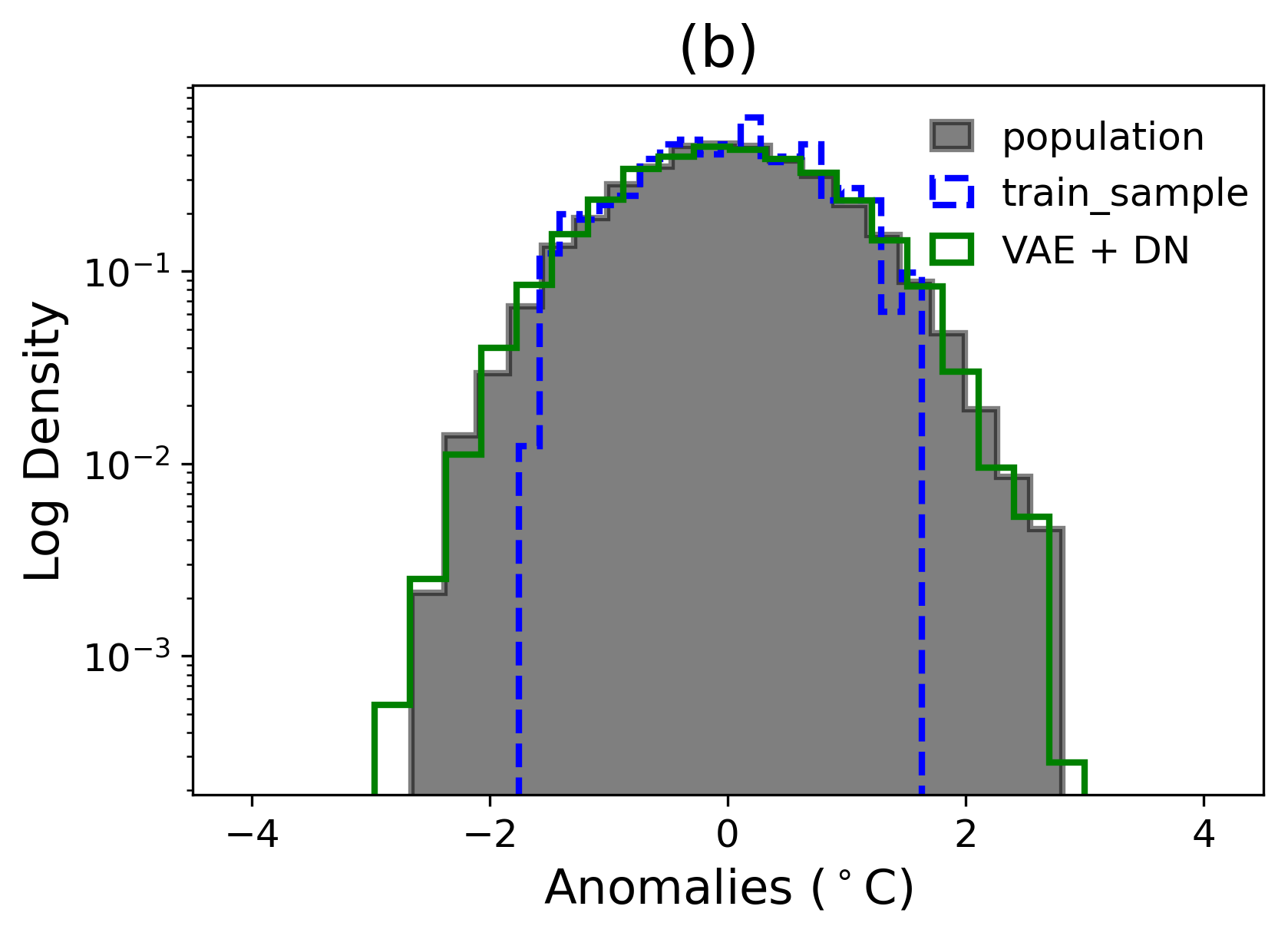}
\end{minipage}
\hfill
\begin{minipage}{0.32\textwidth}
    \centering
    \includegraphics[width=\linewidth]{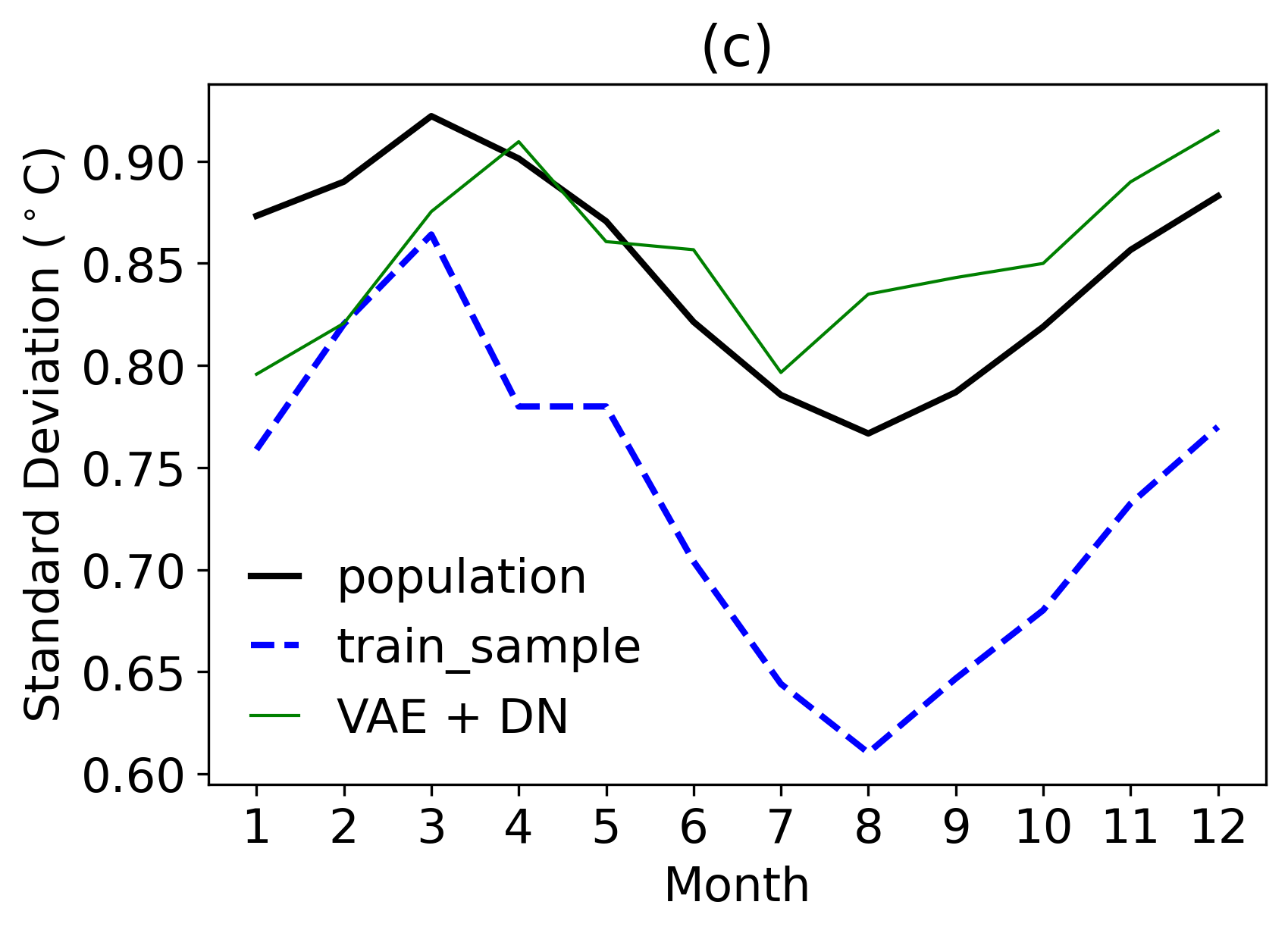}
\end{minipage}

\caption{ a) QQ plots (b) PDF and (c) seasonal cycle of variability in Niño3.4 index for different data products} 
\label{fig:fig4}
\end{figure}

QQ plots of the Niño3.4 index (Fig. \ref{fig:fig4}a) show excellent agreement for the VAE+DN and population percentiles, despite the poor performance in the lower and upper percentiles of the training data. Moreover, Fig. \ref{fig:fig4}b confirms that while the training data fail to represent extreme ENSO events, as indicated by the tails of the PDFs, the boosted ensemble correctly encompasses the variability and extremes in the population. This is also apparent in Figure \ref{fig:fig4}c showing the seasonal cycle for the standard deviation of the Niño3.4 index. Despite the underestimated variability in the training sample, the boosted ensembles largely reproduces the seasonal cycle of the population, although there are slight disagreements, notably in the timing of maximum and minimum variance. This behavior is not surprising, since VAE+DN does not model the time dependence explicitly but infers it from the condition, and the input is standardized using seasonal statistics calculated from limited training data. It is consistent, for instance, with the seasonal cycle for TAS anomalies in the four locations discussed in section \ref{subsec:regions} (Fig. S5 of the supplements). We expect approaches that take into account time dependence, such as VAE + LSTM as in \cite{sorensen2024probabilisticframeworklearningnonintrusive}, to improve the time dependence.

\begin{figure}

% ---------- Row 1 ----------
\begin{minipage}{0.48\textwidth}
    \centering
    \includegraphics[width=\linewidth]{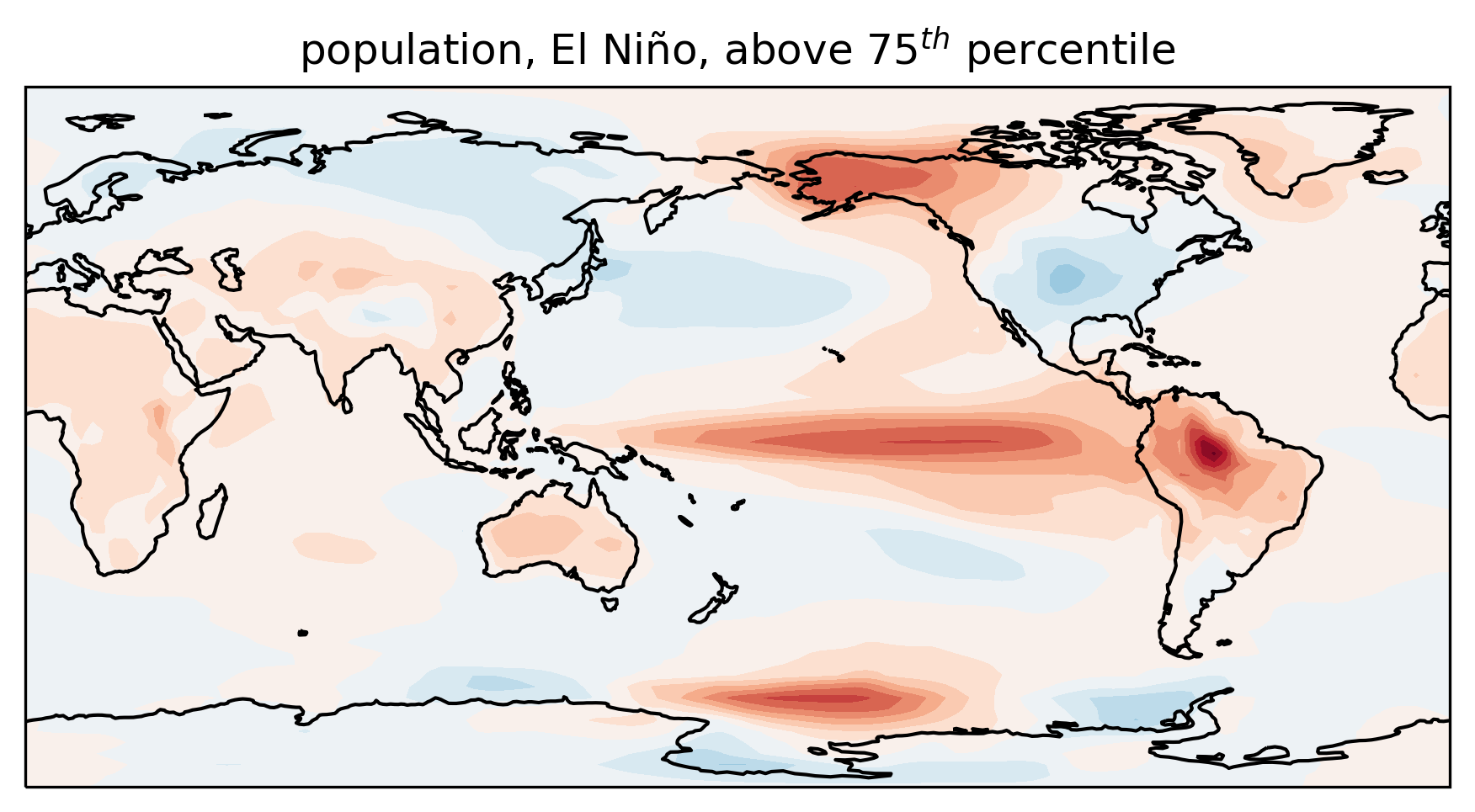}
\end{minipage}\hfill
\begin{minipage}{0.48\textwidth}
    \centering
    \includegraphics[width=\linewidth]{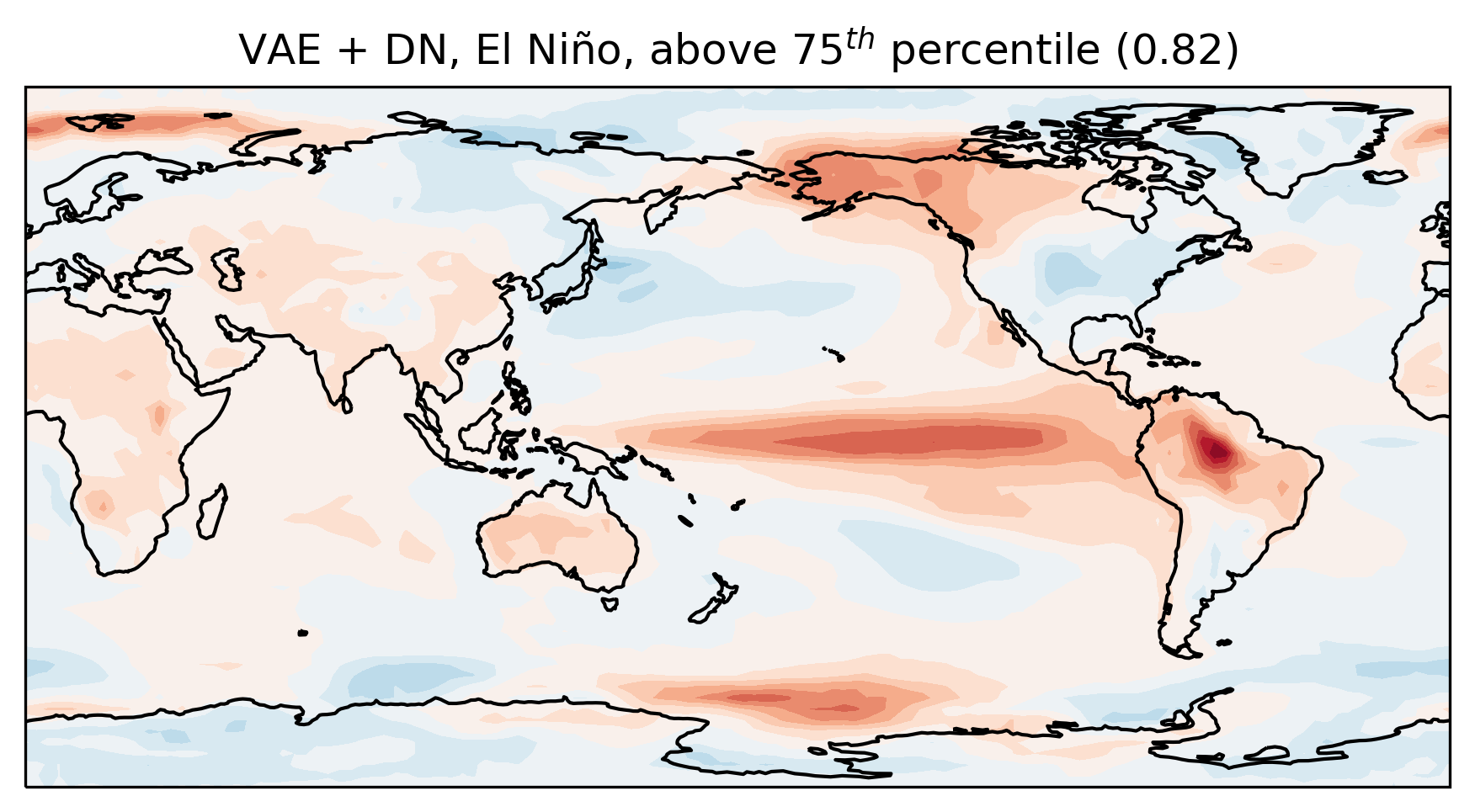}
\end{minipage}

\vspace{0.6em}

% ---------- Row 2 ----------
\begin{minipage}{0.48\textwidth}
    \centering
    \includegraphics[width=\linewidth]{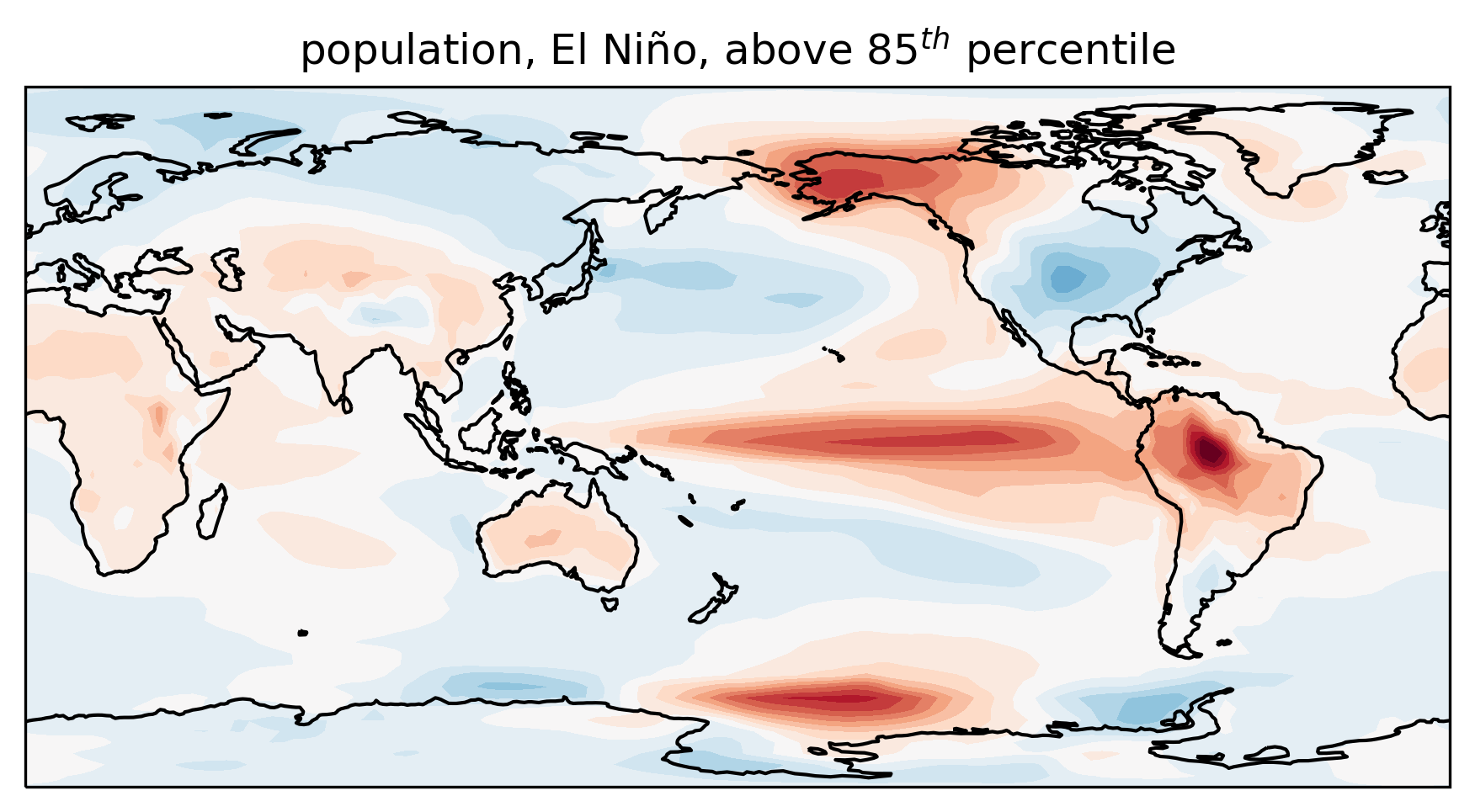}
\end{minipage}\hfill
\begin{minipage}{0.48\textwidth}
    \centering
    \includegraphics[width=\linewidth]{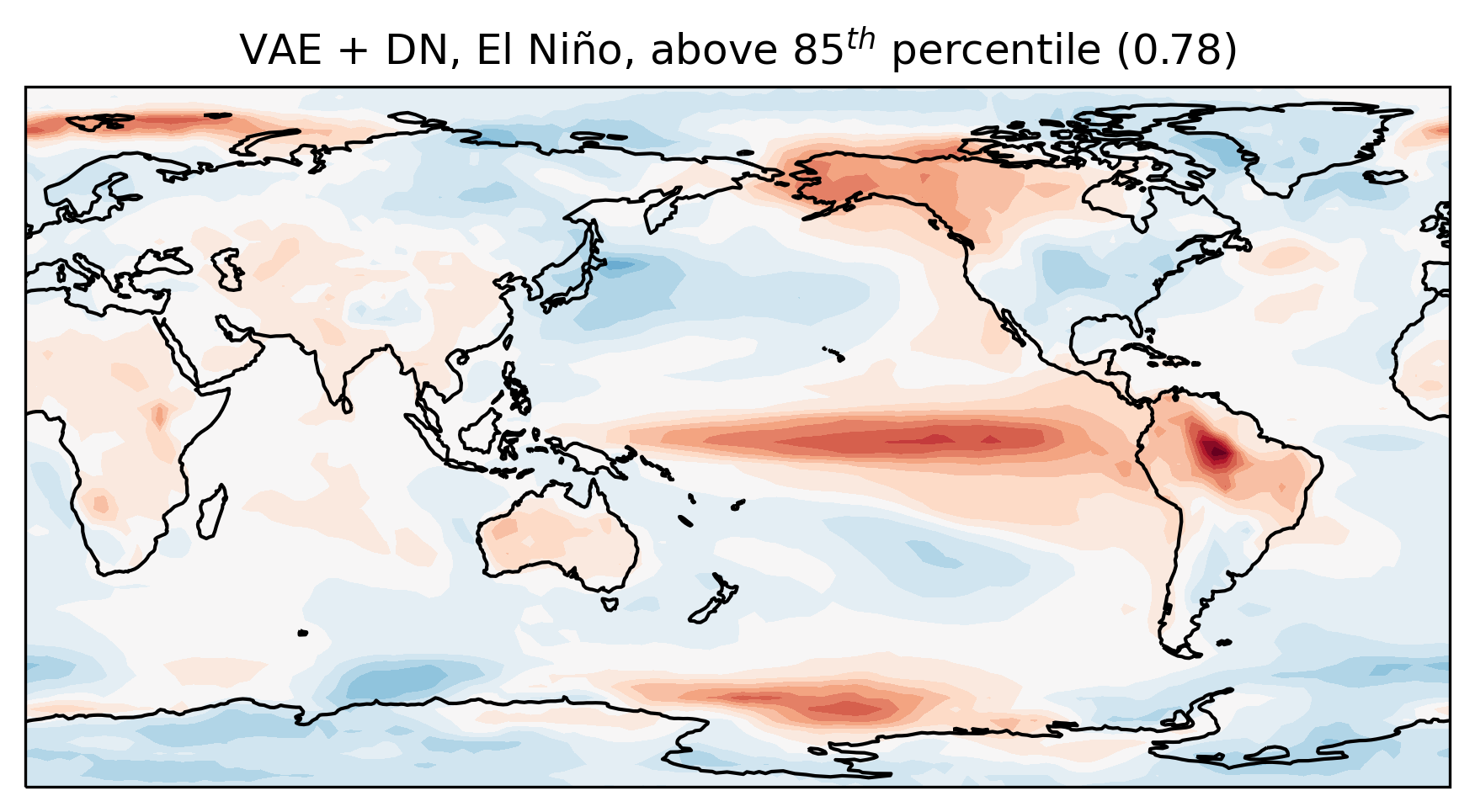}
\end{minipage}

\vspace{0.6em}

% ---------- Row 3 ----------
\begin{minipage}{0.48\textwidth}
    \centering
    \includegraphics[width=\linewidth]{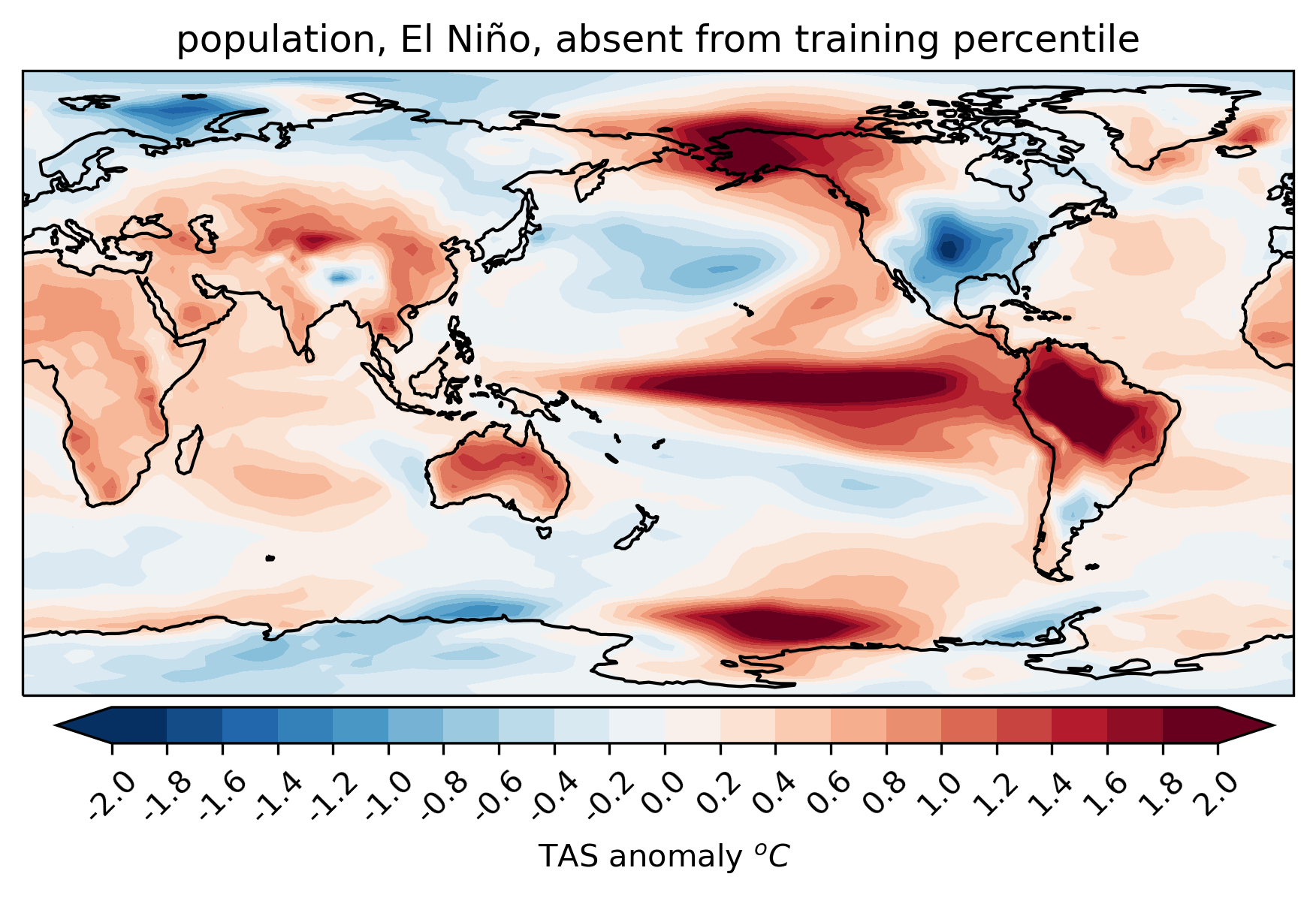}
\end{minipage}\hfill
\begin{minipage}{0.48\textwidth}
    \centering
    \adjustbox{padding=0pt 22pt 0pt 0pt}{\includegraphics[width=\linewidth]{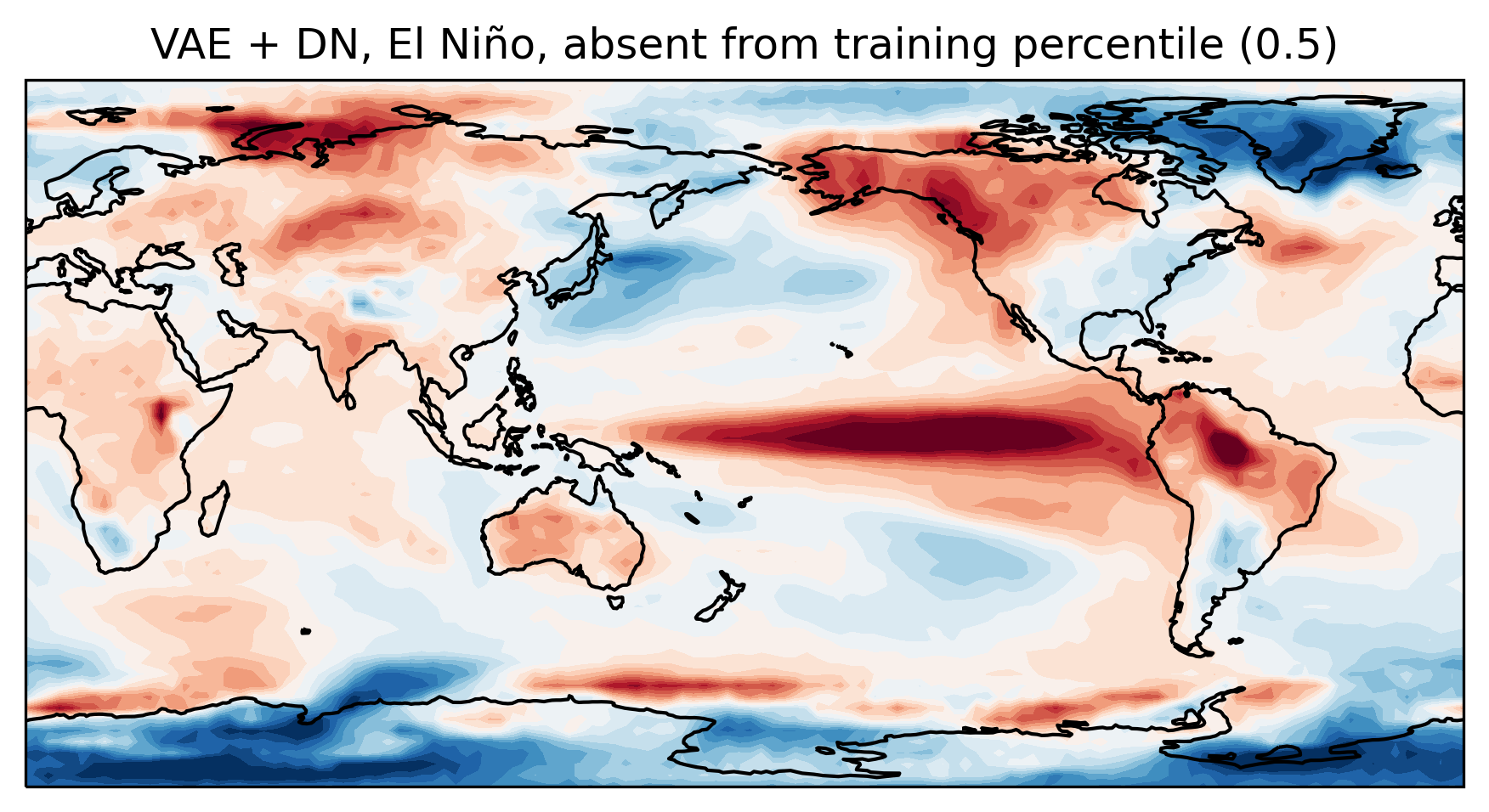}}

\end{minipage}

\caption{Composites of the linearly detrended monthly TAS anomalies for El-Niño events classified using the $75^{th}$, $85^{th}$ and absent from training percentiles based on the Niño3.4 index (rows). First column is the population dataset and the second column is the boosted ensemble from VAE+DN model.} 
\label{fig:fig5}
\end{figure}

Having demonstrated that the VAE+DN model produces TAS fields with a meaningful Niño3.4 index beyond the training sample, we evaluate whether known ENSO teleconnection patterns are present in the boosted ensemble. Figure \ref{fig:fig5} compares the composite maps obtained by averaging the TAS anomalies for three categories of El Niño events, defined as the Niño3.4 index above the percentiles $75^{th}$ and $85^{th}$, and above the highest value in the training data. We compute the percentiles by pooling data from all months, with the latter category representing extreme ENSO events in the boosted ensemble that are absent from the training data. Prior to this computation, we linearly detrend the ensemble members at each month to emphasize interannual variations relevant to ENSO. The rationale is that strong El Niño events drive TAS anomalies in teleconnected regions that on average follow known warming and cooling patterns. For example, strong El Niño episodes, which are characterized by high temperatures in the tropical Pacific, tend to warm western Canada and Alaska, Southeast Asia, Southeast Africa, parts of Australia, and tropical regions of South America, and cool the southern United States and eastern portions of the extratopics in South America. If such patterns are well represented in the CanESM5 raw ensemble, then a physically meaningful generated ensemble trained on CanESM5 data must also represent those patterns.

As Fig. \ref{fig:fig5} shows, the generated TAS patterns agree very well with those of the population, particularly in the regions of known teleconnections listed above. There is no expectation of agreement outside those regions, as the variability there is not necessarily linked to ENSO. The cold anomalies over southern United States appear to be slightly reduced in the boosted ensemble, especially for the absent-from-training category, possibly due to underepresentation of very extreme El Niño events for the linearly detrended index (Fig. S7). Same conclusions hold for La Niña events (Fig. S8) with closely matching patterns of teleconnections but stronger magnitudes of anomalies consistent with overestimation of extreme La Niña events in the linearly detrended index (Fig. S7). The disagreements are mainly the result of biases associated with the small training sample, and the condition embedding coming from a single training realization. In the case of climate projections, we expect improvements for a conditioning that better isolates external forcing, for example, as obtained from the ensemble average of two or more members. The use of a single realization is in line with applications to predictions, since it is desirable to reproduce the phasing of the predictable internal modes of climate variability, such as ENSO.

Finally, to further assess the global structure and physical realism of the generated ensemble, we examine the radially averaged spectral power of the full TAS fields (not anomalies) across all realizations, averaged over DJF and JJA seasons during 1980-2020. Figure S9 shows a close correspondence to the population indicating the that cVAE model is able to reproduce well the variability across spatial scales, the exception being scales below $\sim200$ kms for which there is a small positive bias (consistent with Fig. 1 for TAS anomalies). Those spatial scales are on the order of CanESM5 resolution in and near the tropics. Overall, these results demonstrate that the generated samples reproduce physically meaningful events consistent with the underlying climate. Moreover, they show that the cVAE model is not simply memorizing the input data but learns the underlying climate distribution and is able to produce out-of-distribution samples, even when trained on an extremely limited dataset.

\section{Summary and conclusions} % or: Discussion and conclusions
\label{sec:conclusions}

We employed a conditional Variational Autoencoder, a class of generative machine learning models, to boost the ensemble size of climate simulations given only a single realization of a reference dataset. The work is motivated by the computational limitations of climate models and the resulting trade-off between increasing the ensemble size or the models spatial resolution. Sufficiently large ensembles of climate predictions and projections are critical to inform policy in the face of climate change and the changing frequency of extremes. A large ensemble would help quantifying uncertainty, the likelihood of extreme, the occurrence of low-probability and high-impact events, and to support long-term decision-making. This is one avenue where generative ML can be effective and provide solutions. 

Generative models offer powerful means to learn the underlying distribution of high-dimensional climate fields, making them attractive tools for enhancing climate simulation ensembles. However, the strong non-stationarity of climate data, combined with their limited availability, typically at monthly frequency, present challenges to apply generative models without adaptations. A workaround is to use climate emulators trained on simulations produced with physics-based climate models. In this study, we propose an alternative approach in which we expand ensembles of climate simulations in post-processing by training a generative model with as little as one ensemble member. We demonstrated that even with such a small training sample and with relatively simple model architectures, we can generate large ensembles of physically meaningful simulations. We emphasize that our goal is not to develop the most skillful generative model but rather to create an accessible, simple framework for large-ensemble generation of physically relevant climate simulations, laying the foundation for future work with more advanced models and techniques.

We increased the ensemble size of historical and SSP2-4.5 TAS simulations  produced with CCCma's Canadian Earth system model CanESM5 contributed to CMIP6. We trained a cVAE with monthly snapshots of TAS fields from a single realization of the CanESM5 ensemble, conditioned on a low-dimensional vector acquired from the same TAS maps. The condition embedding was designed to capture the dominant modes of (predictable) variability in the data, including those arising from external forcing.
The trained cVAE was then used at inference time to generate a large TAS field ensemble by sampling the model's latent distribution (Eq. \ref{eq:cVAE_updated}). Our approach was validated by comparing the boosted ensemble with the full 24-member CanESM5 ensemble (excluding training data) for the 1980-2020 period. We showed that the cVAE effectively learns the underlying distribution of TAS anomalies accross both regional and global spatial scales, and it generalizes well to produce physically meaningful TAS fields. Its ability to reproduce extremes and well-known ENSO teleconnection patterns in the population beyond its training data further supports the reliability of the boosted ensemble. The conclusions of this work do not change if a different, randomly chosen ensemble member is used for training. This is supported by the close agreement between statistics of the generated ensemble and the full CanESM5 ensemble over the time scale of analysis. We discussed where the variability inherited from the training sample through conditioning might affect the results (e.g. for the detrended ENSO index), but the conclusions remain unaffected. While learning the predictable internal variability from conditioning is desirable for (phase-locked) seasonal-to-decadal predictions, this is suboptimal in the case for estimating long-term uncertainty. For climate projections, we expect that better isolating the forced component in the condition and using more realizations for training will improve the model's ability to estimate long-term uncertainty of the forced trend, especially resulting from slow evolving variability.

% We increased the ensemble size of historical and SSP2-4.5 simulations of TAS  produced by CCCma's CanESM5 that contributed to CMIP6. We trained a cVAE with monthly snapshots of TAS fields from a single realization of the CanESM5 ensemble, conditioned on a low-dimensional condition embedding vector acquired from the same TAS maps. The condition embedding was aimed at extracting dominant modes of variability in the training data (e.g. from external forcing), . This was then used as a conditioning field for generating large ensembles of realizations at inference time by sampling the latent distribution of the cVAE that is trained to learn the conditional distribution of the climate system (Eq. \ref{eq:elbo}). The approach is evaluated by comparing the boosted ensemble to the full 25-member CanESM5 ensemble over 1980-2020 period. We show the skill of the model in learning the distributions of TAS anomalies both on regional scale, and for the dominant mode of large climate variability --ENSO. The analysis show generalization skill of the model for out-of-sample generation including extreme that were absent from the training data. Finally, the analysis of global teleconnections and spectral energy proves that the generated fields have physical realism and are less likely to be model hallucinations. 

The broader objective of this study is to apply this methodology to operational predictions and projections, helping to boost ensemble sizes during post-processing. In doing so, the challenge of underdispersivity in generative machine learning merits special attention to adequately represent forecast uncertainty and extremes. For the approach followed in this study, we point out two sources of underdispersivity in relation to extremes, one being the non-Gaussian structure of the latent space away from the Gaussian prior and another the loss of variability due to improper formulation of decoder noise, for instance, when MSE is used as a reconstruction loss. We discussed approximations to the latent distribution and the covariance noise structure in the decoder, estimated from the training stage, which greatly improve underdispersivity and performance. We showed that understanding the variability in the model output, generated by the decoder, is especially important for accurate representation of the climate system given the presence of complex spatio-temporal variability in the climate data and its unpredictable weather-induced noise component. In fact, we would argue that the issue of spectral loss in ML models could largely be attributed to deep learning models that often implicitly solve a probability density approximation for a Gaussian output distribution. As a result, variability in the output distribution becomes important for the representation of the fine-scale structure, rather than simply training with a loss function that drives the convergence of the output to the mean, in the case of mean square error, or the median, in the case of mean absolute error \cite{Goodfellow-deeplearning}. 

Future studies will use this framework for climate simulations at higher resolutions, multi-variable ensemble generation, boosting ensembles of CMIP6 scenario runs and seasonal-to-decadal predictions, and for observation-based ensembles to assess limits of predictability and probability of extremes. Moreover, the framework proposed in this study does not explicitly model the time dependence. This can be addressed using autoregressive methods or recurrent neural networks. We recommend either modeling decoder variability as a function of the data whenever possible or providing clear guidelines and explanations when using approximations or more advanced ML techniques. Finally, we expect more expressive and optimally tuned architectures to reduce the biases associated with residual structure in decoder noise.

%----------------
% EXAMPLE FIGURES
%
% \begin{figure}
% \includegraphics{example.png}
% \caption{caption}
% \end{figure}
%
% Giving latex a width will help it to scale the figure properly. A simple trick is to use \textwidth. Try this if large figures run off the side of the page.
% \begin{figure}
% \noindent\includegraphics[width=\textwidth]{anothersample.png}
%\caption{caption}
%\label{pngfiguresample}
%\end{figure}
%
%
% If you get an error about an unknown bounding box, try specifying the width and height of the figure with the natwidth and natheight options. This is common when trying to add a PDF figure without pdflatex.
% \begin{figure}
% \noindent\includegraphics[natwidth=800px,natheight=600px]{samplefigure.pdf}
%\caption{caption}
%\label{pdffiguresample}
%\end{figure}
%
%
% PDFLatex does not seem to be able to process EPS figures. You may want to try the epstopdf package.
%

%
% ---------------
% EXAMPLE TABLE
%
% \begin{table}
% \caption{Time of the Transition Between Phase 1 and Phase 2$^{a}$}
% \centering
% \begin{tabular}{l c}
% \hline
%  Run  & Time (min)  \\
% \hline
%   $l1$  & 260   \\
%   $l2$  & 300   \\
%   $l3$  & 340   \\
%   $h1$  & 270   \\
%   $h2$  & 250   \\
%   $h3$  & 380   \\
%   $r1$  & 370   \\
%   $r2$  & 390   \\
% \hline
% \multicolumn{2}{l}{$^{a}$Footnote text here.}
% \end{tabular}
% \end{table}

%%% End of body of article

%%%%%%%%%%%%%%%%%%%%%%%%%%%%%%%%%%%%%%%%%%%%%%%
%
% DATA SECTION and ACKNOWLEDGMENTS
%
%%%%%%%%%%%%%%%%%%%%%%%%%%%%%%%%%%%%%%%%%%%%%%%

\section*{Open Research Section}
\noindent The data from the CanESM5 model \cite{CanESM5_picontrol_CMIP6} used in the study are publicly available through Earth System Grid Federation (ESGF) data storage servers such as \url{https://aims2.llnl.gov/search/cmip6/?mip_era=CMIP6&activity_id=CMIP&institution_id=CCCma&source_id=CanESM5-1&experiment_id=piControl}. Python codes for training the cVAE models, generating outputs and manuscript analysis/figures can be accessed through \url{https://github.com/ParsaGooya/CCCma_boostENS_JGRMLComp}. All other inquiries should be directed to P. Gooya. (Publication ready and cleaned-up version of codes are being prepared and will be available before final publication.)  
\section*{Conflict of interest}
\noindent The authors declare no conflicts of interest.
\acknowledgments
We thank Dr. Bill Merryfield for his valuable comments on an earlier version of the paper.

\bibliography{References}

@Article{Gooya-2023,
AUTHOR = {Gooya, P. and Swart, N. C. and Hamme, R. C.},
TITLE = {Time-varying changes and uncertainties in the CMIP6~ocean carbon sink from
global to local scale},
JOURNAL = {Earth System Dynamics},
VOLUME = {14},
YEAR = {2023},
NUMBER = {2},
PAGES = {383--398},
URL = {https://esd.copernicus.org/articles/14/383/2023/},
DOI = {10.5194/esd-14-383-2023}
}

@Article{Lehner-2020,
AUTHOR = {Lehner, F. and Deser, C. and Maher, N. and Marotzke, J. and Fischer, E. M. and Brunner, L. and Knutti, R. and Hawkins, E.},
TITLE = {Partitioning climate projection uncertainty with multiple large ensembles and CMIP5/6},
JOURNAL = {Earth System Dynamics},
VOLUME = {11},
YEAR = {2020},
NUMBER = {2},
PAGES = {491--508},
URL = {https://esd.copernicus.org/articles/11/491/2020/},
DOI = {10.5194/esd-11-491-2020}
}

@Article{Swart-2019,
AUTHOR = {Swart, N. C. and Cole, J. N. S. and Kharin, V. V. and Lazare, M. and Scinocca, J. F. and Gillett, N. P. and Anstey, J. and Arora, V. and Christian, J. R. and Hanna, S. and Jiao, Y. and Lee, W. G. and Majaess, F. and Saenko, O. A. and Seiler, C. and Seinen, C. and Shao, A. and Sigmond, M. and Solheim, L. and von Salzen, K. and Yang, D. and Winter, B.},
TITLE = {The Canadian Earth System Model version 5 (CanESM5.0.3)},
JOURNAL = {Geoscientific Model Development},
VOLUME = {12},
YEAR = {2019},
NUMBER = {11},
PAGES = {4823--4873},
URL = {https://gmd.copernicus.org/articles/12/4823/2019/},
DOI = {10.5194/gmd-12-4823-2019}
}

@Article{Sospedra-2021,
AUTHOR = {Sospedra-Alfonso, R. and Merryfield, W. J. and Boer, G. J. and Kharin, V. V. and Lee, W.-S. and Seiler, C. and Christian, J. R.},
TITLE = {Decadal climate predictions with the Canadian Earth System Model version 5 (CanESM5)},
JOURNAL = {Geoscientific Model Development},
VOLUME = {14},
YEAR = {2021},
NUMBER = {11},
PAGES = {6863--6891},
URL = {https://gmd.copernicus.org/articles/14/6863/2021/},
DOI = {10.5194/gmd-14-6863-2021}
}

@article{
diffensemblegen2024,
author = {Lizao Li  and Robert Carver  and Ignacio Lopez-Gomez  and Fei Sha  and John Anderson },
title = {Generative emulation of weather forecast ensembles with diffusion models},
journal = {Science Advances},
volume = {10},
number = {13},
pages = {eadk4489},
year = {2024},
doi = {10.1126/sciadv.adk4489},
URL = {https://www.science.org/doi/abs/10.1126/sciadv.adk4489},
eprint = {https://www.science.org/doi/pdf/10.1126/sciadv.adk4489}}

@article{MLclimate2024,
   author = "Lai, Ching-Yao and Hassanzadeh, Pedram and Sheshadri, Aditi and Sonnewald, Maike and Ferrari, Raffaele and Balaji, Venkatramani",
   title = "Machine Learning for Climate Physics and Simulations", 
   journal= "Annual Review of Condensed Matter Physics",
   year = "2025",
   volume = "16",
   number = "Volume 16, 2025",
   pages = "343-365",
   doi = "https://doi.org/10.1146/annurev-conmatphys-043024-114758",
   url = "https://www.annualreviews.org/content/journals/10.1146/annurev-conmatphys-043024-114758",
   publisher = "Annual Reviews",
   issn = "1947-5462",
   type = "Journal Article",
  }

@article{sacco:hal-03685523,
  TITLE = {{Evaluation of Machine Learning Techniques for Forecast Uncertainty Quantification}},
  AUTHOR = {Sacco, Maximiliano and Ruiz, Juan and Pulido, Manuel and Tandeo, Pierre},
  URL = {https://imt-atlantique.hal.science/hal-03685523},
  JOURNAL = {{Quarterly Journal of the Royal Meteorological Society}},
  PUBLISHER = {{Wiley}},
  VOLUME = {148},
  NUMBER = {749},
  PAGES = {3470-3490},
  YEAR = {2022},
  MONTH = Aug,
  DOI = {10.1002/qj.4362},
  HAL_ID = {hal-03685523},
  HAL_VERSION = {v1},
}

@article{DLpost-processing,
author = {Grönquist, Peter  and Yao, Chengyuan  and Ben-Nun, Tal  and Dryden, Nikoli  and Dueben, Peter  and Li, Shigang  and Hoefler, Torsten },
title = {Deep learning for post-processing ensemble weather forecasts},
journal = {Philosophical Transactions of the Royal Society A: Mathematical, Physical and Engineering Sciences},
volume = {379},
number = {2194},
pages = {20200092},
year = {2021},
doi = {10.1098/rsta.2020.0092},

URL = {https://royalsocietypublishing.org/doi/abs/10.1098/rsta.2020.0092},
eprint = {https://royalsocietypublishing.org/doi/pdf/10.1098/rsta.2020.0092}
}

@article{simultaneous2025,
  title={Simultaneous emulation and downscaling with physically-consistent deep learning-based regional ocean emulators},
  author={Lupin-Jimenez, Leonard and Darman, Moein and Hazarika, Subhashis and Wu, Tianning and Gray, Michael and He, Ruyoing and Wong, Anthony and Chattopadhyay, Ashesh},
  journal={arXiv preprint arXiv:2501.05058},
  year={2025},
  doi={10.48550/arXiv.2501.05058}
}

@article{cachay2024probabilistic,
  title={Probabilistic Emulation of a Global Climate Model with Spherical DYffusion},
  author={R{\"u}hling Cachay, Salva and Henn, Brian and Watt-Meyer, Oliver and Bretherton, Christopher S. and Yu, Rose},
  journal={arXiv preprint arXiv:2406.14798v2},
  year={2024},
  doi={10.48550/arXiv.2406.14798},
  note={NeurIPS 2024 submission; code is available at the provided URL.}
}

@article{dynamicalgeneariveds2024,
  title={Dynamical-generative downscaling of climate model ensembles},
  author={Lopez-Gomez, Ignacio and Wan, Zhong Yi and Zepeda-Núñez, Leonardo and Schneider, Tapio and Anderson, John and Sha, Fei},
  journal={arXiv preprint arXiv:2410.01776v1},
  year={2024},
  doi={10.48550/arXiv.2410.01776}
}

@article{daust2024capturing,
  title={Capturing Climatic Variability: Using Deep Learning for Stochastic Downscaling},
  author={Daust, Kiri and Monahan, Adam},
  journal={arXiv preprint arXiv:2406.02587},
  year={2024},
  doi={10.48550/arXiv.2406.02587},
  note={Submitted to Artificial Intelligence for the Earth Systems AMS Journal}
}

@misc{kingma2014,
      title={Auto-Encoding Variational Bayes}, 
      author={Diederik P Kingma and Max Welling},
      year={2022},
      eprint={1312.6114},
      archivePrefix={arXiv},
      primaryClass={stat.ML},
      url={https://arxiv.org/abs/1312.6114}, 
}

@article{rezende2014,
  title={Stochastic Backpropagation and Approximate Inference in Deep Generative Models},
  author={Rezende, Danilo Jimenez and Mohamed, Shakir and Wierstra, Daan},
  booktitle={Proceedings of the 31st International Conference on Machine Learning},
  editor={Xing, Eric P. and Jebara, Tony},
  series={Proceedings of Machine Learning Research},
  volume={32},
  number={2},
  pages={1278--1286},
  year={2014},
  address={Beijing, China},
  month={22--24 Jun},
  publisher={PMLR},
  url={https://proceedings.mlr.press/v32/rezende14.html}
}

@misc{ghosh2020,
      title={From Variational to Deterministic Autoencoders}, 
      author={Partha Ghosh and Mehdi S. M. Sajjadi and Antonio Vergari and Michael Black and Bernhard Schölkopf},
      year={2020},
      eprint={1903.12436},
      archivePrefix={arXiv},
      primaryClass={cs.LG},
      url={https://arxiv.org/abs/1903.12436}, 
}

@article{szwarcman2024,
  title={Quantizing reconstruction losses for improving weather data synthesis},
  author={Szwarcman, Daniela and Guevara, Jorge and Macedo, Maysa M. G. and Zadrozny, Bianca and Watson, Campbell and Rosa, Laura and Oliveira, Dario A. B.},
  journal={Scientific Reports},
  volume={14},
  number={1},
  pages={3396},
  year={2024},
  month={Feb 9},
  doi={10.1038/s41598-024-52773-2},
  url={https://doi.org/10.1038/s41598-024-52773-2}
}

@article{cVAE2015,
 author = {Sohn, Kihyuk and Lee, Honglak and Yan, Xinchen},
 booktitle = {Advances in Neural Information Processing Systems},
 editor = {C. Cortes and N. Lawrence and D. Lee and M. Sugiyama and R. Garnett},
 pages = {},
 publisher = {Curran Associates, Inc.},
 title = {Learning Structured Output Representation using Deep Conditional Generative Models},
 url = {https://proceedings.neurips.cc/paper_files/paper/2015/file/8d55a249e6baa5c06772297520da2051-Paper.pdf},
 volume = {28},
 year = {2015}
}

@article{Leutbecher2019,
  author = {Martin Leutbecher},
  title = {Ensemble size: How suboptimal is less than infinity?},
  journal = {Quarterly Journal of the Royal Meteorological Society},
  volume = {145},
  number = {S1},
  pages = {107-128},
  year = {2019},
  doi = {10.1002/qj.3387},
  url = {https://rmets.onlinelibrary.wiley.com/doi/abs/10.1002/qj.3387},
  eprint = {https://rmets.onlinelibrary.wiley.com/doi/pdf/10.1002/qj.3387}
}

@article{distVAE2023,
 author = {An, Seunghwan and Jeon, Jong-June},
 booktitle = {Advances in Neural Information Processing Systems},
 editor = {A. Oh and T. Naumann and A. Globerson and K. Saenko and M. Hardt and S. Levine},
 pages = {57825--57851},
 publisher = {Curran Associates, Inc.},
 title = {Distributional Learning of Variational AutoEncoder: Application to Synthetic Data Generation},
 url = {https://proceedings.neurips.cc/paper_files/paper/2023/file/b456a00e145ad56f6f251f79f8c8a7de-Paper-Conference.pdf},
 volume = {36},
 year = {2023}
}

@article{IEEE,
  author={Oliveira, Dario A. B. and Diaz, Jorge G. and Zadrozny, Bianca and Watson, Campbell D. and Zhu, Xiao Xiang},
  booktitle={IGARSS 2022 - 2022 IEEE International Geoscience and Remote Sensing Symposium}, 
  title={Controlling Weather Field Synthesis Using Variational Autoencoders}, 
  year={2022},
  volume={},
  number={},
  pages={5027-5030},
  doi={10.1109/IGARSS46834.2022.9884668}}

@article{Taiwan,
author = {Hsieh, Min-Ken and Wu, Chien-Ming},
title = {Developing an Explainable Variational Autoencoder (VAE) Framework for Accurate Representation of Local Circulation in Taiwan},
journal = {Journal of Geophysical Research: Atmospheres},
volume = {129},
number = {12},
pages = {e2024JD041167},
doi = {https://doi.org/10.1029/2024JD041167},
url = {https://agupubs.onlinelibrary.wiley.com/doi/abs/10.1029/2024JD041167},
eprint = {https://agupubs.onlinelibrary.wiley.com/doi/pdf/10.1029/2024JD041167},
note = {e2024JD041167 2024JD041167},
year = {2024}
}

@article{AtmConvVAE2021,
author = {Mooers, Griffin and Tuyls, Jens and Mandt, Stephan and Pritchard, Mike and Beucler, Tom G},
title = {Generative Modeling of Atmospheric Convection},
year = {2021},
isbn = {9781450388481},
publisher = {Association for Computing Machinery},
address = {New York, NY, USA},
url = {https://doi.org/10.1145/3429309.3429324},
doi = {10.1145/3429309.3429324},
booktitle = {Proceedings of the 10th International Conference on Climate Informatics},
pages = {98–105},
numpages = {8},
location = {virtual, United Kingdom},
series = {CI2020}
}

@article{dhariwal2021diffusion,
  title     = {Diffusion Models Beat {GANs} on Image Synthesis},
  author    = {Dhariwal, Prafulla and Nichol, Alex},
  booktitle = {Advances in Neural Information Processing Systems},
  volume    = {34},
  pages     = {8780--8794},
  year      = {2021}
}

@misc{ho2022imagenvideodiffusion,
  title       = {Imagen Video: High Definition Video Generation with Diffusion Models},
  author      = {Ho, J. and Chan, W. and Saharia, C. and Whang, J. and Gao, R. and Gritsenko, A. and Kingma, D. P. and Poole, B. and Norouzi, M. and Fleet, D. J. and Salimans, T.},
  howpublished = {arXiv preprint \texttt{arXiv:2210.02303}},
  year        = {2022},
  month       = oct,
  note        = {arXiv:2210.02303 [cs.CV], 5 October 2022}
}

@article{rombach2022highdiffusion,
  title       = {High-Resolution Image Synthesis with Latent Diffusion Models},
  author      = {Rombach, Robin and Blattmann, Andreas and Lorenz, Dominik and Esser, Patrick and Ommer, Bj{\"o}rn},
  booktitle   = {Proceedings of the IEEE Conference on Computer Vision and Pattern Recognition (CVPR)},
  pages       = {10684--10695},
  year        = {2022},
  organization = {IEEE},
  url         = {https://github.com/CompVis/latent-diffusion}
}

@article{gencast,
  title        = {Probabilistic weather forecasting with machine learning},
  author       = {Price, Ilan and Sanchez-Gonzalez, Alvaro and Alet, Ferran and Andersson, Tom R. and El-Kadi, Andrew and Masters, Dominic and Ewalds, Timo and Stott, Jacklynn and Mohamed, Shakir and Battaglia, Peter and Lam, Remi and Willson, Matthew},
  journal      = {Nature},
  year         = {2025},
  volume       = {637},
  number       = {8044},
  pages        = {84--90},
  doi          = {10.1038/s41586-024-08252-9},
  url          = {https://doi.org/10.1038/s41586-024-08252-9}
}

@article{GANS2014Goodfellow,
 author = {Goodfellow, Ian and Pouget-Abadie, Jean and Mirza, Mehdi and Xu, Bing and Warde-Farley, David and Ozair, Sherjil and Courville, Aaron and Bengio, Yoshua},
 booktitle = {Advances in Neural Information Processing Systems},
 editor = {Z. Ghahramani and M. Welling and C. Cortes and N. Lawrence and K.Q. Weinberger},
 pages = {},
 publisher = {Curran Associates, Inc.},
 title = {Generative Adversarial Nets},
 url = {https://proceedings.neurips.cc/paper_files/paper/2014/file/5ca3e9b122f61f8f06494c97b1afccf3-Paper.pdf},
 volume = {27},
 year = {2014}
}

@book{prince2023understanding,
        author = "Simon J.D. Prince",
        title = "Understanding Deep Learning",
        publisher = "The MIT Press",
        year = 2023,
        url = "http://udlbook.com"
    }

@misc{dorta2018structureduncertaintypredictionnetworks,
      title={Structured Uncertainty Prediction Networks}, 
      author={Garoe Dorta and Sara Vicente and Lourdes Agapito and Neill D. F. Campbell and Ivor Simpson},
      year={2018},
      eprint={1802.07079},
      archivePrefix={arXiv},
      primaryClass={stat.ML},
      url={https://arxiv.org/abs/1802.07079}, 
}

@misc{betaannealing,
      title={$\beta$-Annealed Variational Autoencoder for glitches}, 
      author={Sivaramakrishnan Sankarapandian and Brian Kulis},
      year={2021},
      eprint={2107.10667},
      archivePrefix={arXiv},
      primaryClass={cs.LG},
      url={https://arxiv.org/abs/2107.10667}, 
}

@Article{eyering-2016,
AUTHOR = {Eyring, V. and Bony, S. and Meehl, G. A. and Senior, C. A. and Stevens, B. and Stouffer, R. J. and Taylor, K. E.},
TITLE = {Overview of the Coupled Model Intercomparison Project Phase 6 (CMIP6)
experimental design and organization},
JOURNAL = {Geoscientific Model Development},
VOLUME = {9},
YEAR = {2016},
NUMBER = {5},
PAGES = {1937--1958},
URL = {https://gmd.copernicus.org/articles/9/1937/2016/},
DOI = {10.5194/gmd-9-1937-2016}
}

@misc{doublepenalty,
      title={Fixing the Double Penalty in Data-Driven Weather Forecasting Through a Modified Spherical Harmonic Loss Function}, 
      author={Christopher Subich and Syed Zahid Husain and Leo Separovic and Jing Yang},
      year={2025},
      eprint={2501.19374},
      archivePrefix={arXiv},
      primaryClass={cs.LG},
      url={https://arxiv.org/abs/2501.19374}, 
}

@book{Goodfellow-deeplearning,
    title={Deep Learning},
    author={Ian Goodfellow and Yoshua Bengio and Aaron Courville},
    publisher={MIT Press},
    note={\url{http://www.deeplearningbook.org}},
    year={2016}
}

@misc{CanESM5_picontrol_CMIP6,
      url = {https://doi.org/10.22033/ESGF/CMIP6.17424},
      title = {CCCma CanESM5.1 model output prepared for CMIP6 CMIP piControl},
      publisher = {Earth System Grid Federation},
      year = {2019},
      author = {Swart, Neil Cameron and Cole, Jason N.S. and Kharin, Viatcheslav V. and Lazare, Mike and Scinocca, John F. and Gillett, Nathan P. and Anstey, James and Arora, Vivek and Christian, James R. and Jiao, Yanjun and Lee, Warren G. and Majaess, Fouad and Saenko, Oleg A. and Seiler, Christian and Seinen, Clint and Shao, Andrew and Solheim, Larry and von Salzen, Knut and Yang, Duo and Winter, Barbara and Sigmond, Michael and Abraham, Carsten and Akingunola, Ayodeji and Reader, Catherine},
      doi = {10.22033/ESGF/CMIP6.17424}
}

@article{non-intrusice-correction-det,
author = {Barthel Sorensen, Benedikt and Charalampopoulos, A. and Zhang, S. and Harrop, B. E. and Leung, L. R. and Sapsis, T. P.},
title = {A Non-Intrusive Machine Learning Framework for Debiasing Long-Time Coarse Resolution Climate Simulations and Quantifying Rare Events Statistics},
journal = {Journal of Advances in Modeling Earth Systems},
volume = {16},
number = {3},
pages = {e2023MS004122},
doi = {https://doi.org/10.1029/2023MS004122},
url = {https://agupubs.onlinelibrary.wiley.com/doi/abs/10.1029/2023MS004122},
eprint = {https://agupubs.onlinelibrary.wiley.com/doi/pdf/10.1029/2023MS004122},
note = {e2023MS004122 2023MS004122},
year = {2024}
}

@misc{sorensen2024probabilisticframeworklearningnonintrusive,
      title={A probabilistic framework for learning non-intrusive corrections to long-time climate simulations from short-time training data}, 
      author={Benedikt Barthel Sorensen and Leonardo Zepeda-Núñez and Ignacio Lopez-Gomez and Zhong Yi Wan and Rob Carver and Fei Sha and Themistoklis Sapsis},
      year={2024},
      eprint={2408.02688},
      archivePrefix={arXiv},
      primaryClass={cs.LG},
      url={https://arxiv.org/abs/2408.02688}, 
}

@misc{wang2025gen2generativepredictioncorrectionframework,
      title={GEN2: A Generative Prediction-Correction Framework for Long-time Emulations of Spatially-Resolved Climate Extremes}, 
      author={Mengze Wang and Benedikt Barthel Sorensen and Themistoklis Sapsis},
      year={2025},
      eprint={2508.15196},
      archivePrefix={arXiv},
      primaryClass={physics.comp-ph},
      url={https://arxiv.org/abs/2508.15196}, 
}

@article{nbsgf17,
author = {Daniele Nerini and Nikola Besic and Ioannis Sideris and Urs Germann and Loris Foresti},
title = {A non-stationary stochastic ensemble generator for radar rainfall fields based on the short-space Fourier transform},
journal = {Hydrol. Earth Syst. Sci.},
volume = {21},
number = {},
pages = {2777--2797},
OPTkeywords = {},
doi = {10.5194/hess-21-2777-2017},
url = {https://doi.org/10.5194/hess-21-2777-2017},
OPTeprint = {},
OPTnote = {},
OPTabstract = {},
year = {2017}
}

@inproceedings{gooya2025probabilisticbiasadjustmentseasonal,
  title={Probabilistic bias adjustment of seasonal predictions of Arctic Sea Ice Concentration},
  author={Gooya, Parsa and Sospedra-Alfonso, Reinel},
  booktitle={NeurIPS 2025 Workshop on Tackling Climate Change with Machine Learning},
  url={https://www.climatechange.ai/papers/neurips2025/87},
  year={2025}
}

@article{merryfield2020,
author = {William J. Merryfield and Johanna Baehr and Lauriane Batté and  Emily J. Becker and  Amy H. Butler and  Caio A. S. Coelho and Gokhan Danabasoglu and Paul A. Dirmeyer and Francisco J. Doblas-Reyes and Daniela I. V. Domeisen and Laura Ferranti and Tatiana Ilynia and Arun Kumar and Wolfgang A. Müller and Michel Rixen and Andrew W. Robertson and Doug M. Smith and Yuhei Takaya and Matthias Tuma and Frederic Vitart and Christopher J. White and Mariano S. Alvarez and Constantin Ardilouze and Hannah Attard and Cory Baggett and Magdalena A. Balmaseda and Asmerom F. Beraki and Partha S. Bhattacharjee and Roberto Bilbao and Felipe M. de Andrade and Michael J. DeFlorio and Leandro B. Díaz and Muhammad Azhar Ehsan and Georgios Fragkoulidis and Alex O. Gonzalez and Sam Grainger and Benjamin W. Green and Momme C. Hell and Johnna M. Infanti and Katharina Isensee and Takahito Kataoka and Ben P. Kirtman and Nicholas P. Klingaman and June-Yi Lee and Kirsten Mayer and Roseanna McKay and Jennifer V. Mecking and Douglas E. Miller and Nele Neddermann and Ching Ho Justin Ng and Albert Ossó and Klaus Pankatz and Simon Peatman and Kathy Pegion and Judith Perlwitz and G. Cristina Recalde-Coronel and Annika Reintges and Christoph Renkl and Balakrishnan Solaraju-Murali and Aaron Spring and Cristiana Stan and Y. Qiang Sun and Carly R. Tozer and Nicolas Vigaud and Steven Woolnough and Stephen Yeager},
title = {Current and emerging developments in Subseasonal to Decadal Prediction},
journal = {Bulletin of the American Meteorological Society},
volume = {101},
number = {6},
pages = {E869--E896},
year = {2020},
doi = {10.1175/BAMS-D-19-0037.1},
URL = {https://doi.org/10.1175/BAMS-D-19-0037.1},
}

@article{bkm13,
author = {Boer, G. J. and Kharin, V. V. and Merryfield, W. J.},
title = {Decadal predictability and forecast skill},
journal = {Climate Dynamics},
volume = {41},
OPTnumber = {},
pages = {1817--1833},
year = {2013},
doi = {10.1007/s00382-013-1705-0},
URL = {https://link.springer.com/article/10.1007/s00382-013-1705-0},
}

@article{sb19,
author = {Reinel Sospedra-Alfonso and George J. Boer},
title = {Assessing the impact of initialization on decadal prediction skill},
journal = {Geophysical Research Letters},
volume = {47},
number = {4},
pages = {e2019GL086361},
year = {2020},
doi = {10.1029/2019GL086361},
URL = {https://doi.org/10.1029/2019GL086361},
}

% \clearpage

% \section*{References From the Supporting Information}

\nocite{GANS2014Goodfellow}
\nocite{Taiwan}
\nocite{dhariwal2021diffusion}
\nocite{rombach2022highdiffusion} 
\nocite{ho2022imagenvideodiffusion}
\nocite{gencast}
\nocite{betaannealing}
% \bibliography{References_SI}

\end{document}